\documentclass[manuscript,screen,nonacm,acmsmall]{acmart}
\AtBeginDocument{%
  }

\usepackage{siunitx}
\usepackage{cleveref}
\usepackage{setspace}
\usepackage{changepage}
\usepackage{subcaption}
\usepackage{tikz}
\usepackage{pgfplots}
\usepackage{pgfplotstable} 
\usetikzlibrary{arrows,matrix,calc,automata,positioning}
\pgfplotsset{compat=1.18} 
\usepgfplotslibrary{statistics} 

\DeclareUnicodeCharacter{2212}{-}
\usepackage{graphicx}
\usepackage{xparse}

\graphicspath{{Emoji_pdf/}}
\DeclareGraphicsExtensions{.pdf,.png,.jpg}

\NewDocumentCommand{\emoji}{ O{1.1} m }{%
  \raisebox{-0.2ex}{\includegraphics[height=#1em]{emoji/#2.png}}%
}

\definecolor{niceblue}{HTML}{4682B4}

\definecolor{nicered}{HTML}{FA8072}

\definecolor{nicegreen}{HTML}{2E8B57}  

\definecolor{niceblueLine}{HTML}{315f7a}
\definecolor{niceredLine}{HTML}{c75f4d}
\definecolor{nicegreenLine}{HTML}{1f5e3c}

\pgfplotscreateplotcyclelist{nicecolors}{%
  {fill=niceblue,  draw=niceblueLine},
  {fill=nicered,   draw=niceredLine},
  {fill=nicegreen, draw=nicegreenLine},
}

\definecolor{groupA}{HTML}{4682B4}   
\definecolor{groupB}{RGB}{94,60,153}    
\definecolor{groupC}{RGB}{35,132,67}    

\usepackage{graphicx} 

\definecolor{groupAfill}{RGB}{33,113,181}   
\definecolor{groupBfill}{RGB}{94,60,153}    
\definecolor{groupCfill}{RGB}{35,132,67}    

\definecolor{groupAline}{RGB}{23,78,125}    
\definecolor{groupBline}{RGB}{63,40,103}    
\definecolor{groupCline}{RGB}{24,92,47}     

\pgfplotscreateplotcyclelist{groupcolors}{%
  {fill=groupAfill, draw=groupAline},
  {fill=groupBfill, draw=groupBline},
  {fill=groupCfill, draw=groupCline},
}

\usepackage{enumitem} 

\usepackage[most]{tcolorbox} 
\tcbset{enhanced, breakable, frame hidden, sharp corners, boxrule=0pt}

\newtcolorbox{userbubble}{colback=gray!10, left=8pt, right=8pt, top=6pt, bottom=6pt,
  borderline west={2pt}{0pt}{gray!60}}
\newtcolorbox{aibubble}{colback=blue!6, left=8pt, right=8pt, top=6pt, bottom=6pt,
  borderline west={2pt}{0pt}{blue!50!black}}

\newcommand{\userlabel}[1]{\textbf{User (#1)}}
\newcommand{\ailabel}{\textbf{AI}}

\begin{document}

\title{Persuasive or Neutral? A Field Experiment on Generative AI in Online Travel Planning}

\renewcommand{\shorttitle}{A Field Experiment on Generative AI in Itinerary Planning}


\author{Lynna Jirpongopas}
\email{ljirpong@gmail.com}
\affiliation{%
  \institution{University of Freiburg}
  \streetaddress{Rempartstr. 16}
  \city{Freiburg}
  \country{Germany}
}

\author{Bernhard Lutz}
\email{bernhard.lutz@is.uni-freiburg.de}
\affiliation{%
  \institution{University of Freiburg}
  \streetaddress{Rempartstr. 16}
  \city{Freiburg}
  \country{Germany}
}
\author{Jörg Ebner}
\email{joerg.ebner@is.uni-freiburg.de}
\affiliation{%
  \institution{University of Freiburg}
  \streetaddress{Rempartstr. 16}
  \city{Freiburg}
  \country{Germany}
}

\author{Rustam Vahidov}
\email{rustam.vahidov@concordia.ca}
\affiliation{%
  \institution{Concordia University}
  \streetaddress{Guy St 1450 }
  \city{Montréal}
  \country{Canada}
}

\author{Dirk Neumann}
\email{dirk.neumann@is.uni-freiburg.de}
\affiliation{%
  \institution{University of Freiburg}
  \streetaddress{Rempartstr. 16}
  \city{Freiburg}
  \country{Germany}
}

\renewcommand{\shortauthors}{Jirpongopas et al.}

\begin{abstract}
Generative AI (GenAI) offers new opportunities for customer support in online travel agencies, yet little is known about how its design influences user engagement, purchase behavior, and user experience. We report results from a randomized field experiment in online travel itinerary planning, comparing GenAI that expressed (A) positive enthusiasm, (B) neutral expression, and (C) no tone instructions (control). Users in group A wrote significantly longer prompts than those in groups B and C. At the same time, users in groups A and B were more likely to purchase subscriptions of the webservice. We further analyze linguistic cues across experimental groups to explore differences in user experience and explain subscription purchases and affiliate link clicks based on these cues. Our findings provide implications for the design of persuasive and engaging GenAI interfaces in consumer-facing contexts and contribute to understanding how linguistic framing shapes user behavior in AI-mediated decision support.
\end{abstract}

\ccsdesc[500]{Information systems~Web services}
\ccsdesc[500]{Information systems~Expert systems}
\ccsdesc[500]{Human-centered computing~Field studies}
\ccsdesc[300]{Human-centered computing~Empirical studies in interaction design}

\keywords{Online travel agency, generative AI, travel itinerary planning, user behavior, user experience, field experiment}


\maketitle

\newcommand{\highlightcue}[1]{\texttt{#1}}

\section{Introduction}
\label{sec:introduction}


Over the past few years, online travel agencies (OTAs) have been continuously replacing traditional travel agencies. In 2023, the global online travel industry generated \$512.5 billion of revenue and is expected to reach 1.26 trillion by 2032. In particular, the vast majority of travelers (72\%) said they preferred to book their trips online, compared to only 12\% that preferred to use a travel agency.\footnote{\url{https://www.travelperk.com/blog/online-travel-booking-statistics/}, last access September 10, 2025.} 


Travel itinerary planning refers to organizing a holiday trip through various cities or towns, including the preferred activities such as visits to national parks, sights, museums, or beaches \citep{Roy.2011,zhang.2016.Trips}. Given a fixed time period and budget, travelers aim to maximize the overall experience of their trip \citep{Zhang.2016.CHI}. Planning an optimal itinerary through various cities with little knowledge about the country or region while accounting for time and budget constraints can be challenging  \citep{Lim.2019}. At the same time, it is emotionally charged, driven by excitement, anticipation, and fear of missing out.





This complexity makes travel itinerary planning a promising domain for GenAI, a natural candidate for integration into OTAs. While GenAI can easily assist customers with simple bookings such as individual flights, train tickets, or rental cars \citep{He.2025}, its potential should be even more pronounced when booking a complex itinerary through various destinations, including the desired activities of the traveler. GenAI not only provides knowledge about most places in the world, including points of interest, sights and possible activities, it can also guide users through the entire booking process. 
Prior research has acknowledged the potential of GenAI in many areas, including planning, marketing, customer-relationship management, and customer support \citep[e.g.,][]{Adam.2021,Feuerriegel.2024,han.2023,Lee.2025,Seymour.2024}. Despite rapid deployment of AI assistants across domains, there is little evidence on how GenAI should be designed for consumer-facing contexts. Benchmark studies on GenAI in itinerary planning demonstrate technical challenges such as hallucinations and constraint handling \citep{Xie.2024,Shen.2025}, while prototype systems show that GenAI can in principle generate itineraries \citep{Ju.2024,Udandarao.2025}. However, it remains unclear how GenAI should communicate with users in real-world planning scenarios to foster engagement and purchase behavior.



A growing body of HCI and service research shows that the tone of AI communication plays a central role in shaping user experience. Warm and enthusiastic messages can increase perceptions of social presence and engagement \citep{Kull.2021}, while neutral or competence-oriented tones may be more effective in tasks that demand accuracy and credibility \citep{Packard.2021,Li.2019}. Language expectancy theory suggests that humans have different normative expectations towards the language that is used in different contexts and social roles. Deviations from these norms can positively or negatively influence the communicator's persuasiveness. For instance, humans generally expect communication in professional working environments to be more objective, whereas communication with friends and family are considered more personal and emotional \citep{Burgoon.1985.LET,Burgoon.1995.LET}. Contextualized to itinerary planning with GenAI, emotionally expressive AI responses may constitute a positive expectancy violation, increasing persuasiveness of the AI, leading to greater user engagement and more frequent purchases. Conversely, users might also expect a more neutral tone, reflecting the rational and objective nature of AI. In fact, prior studies have found limited results regarding how (Gen)AI in customer service should be designed \citep{Gnewuch.2017,han.2023,Zhang.2024}. Recent results by \citet{han.2023} suggest that chatbots in customer support should communicate using more neutral and objective rather than emotional language. 
Against this background, online itinerary planning with GenAI represents a distinct use case where users may seek not only efficiency but also excitement and personalization--needs that require effective communication strategies. 


The tone used by GenAI in customer service influences user experience  \citep{Hu.2018.CHI,Okoso.2025}. Different user experiences are then reflected in the way humans communicate \citep{Fang.2025,Ferland.2025,He.2024}. \citet{Jang.2017} demonstrated that UX elements can be extracted from user-generated text using ``Linguistic Inquiry and Word Count'' \citep[LIWC,][]{Pennebaker.2015} and that these linguistic cues significantly affect user satisfaction. Therefore, we also perform an explorative analysis of the usage of linguistic cues in users' input prompts. In particular, we aim to explain the subsequent purchase behavior across the experimental groups based on these cues.


We present results from a randomized field experiment, conducted at a US startup that offers a travel itinerary planning service based on OpenAI GPT models. A total of 282 users participated in the A/B/C test from November 3, 2024 to April 14, 2025. We added tone instructions in the base prompt of the chatbot to be either (A)~positive enthusiastic, using emoji when appropriate, (B)~neutral,  expressionless, or (C)~no tone instruction given. 
Our findings suggest that users in Group A engaged more with GenAI, measured as the length of input prompts, than users in Groups B and C. In addition, we find  that users of Group A and B purchased more subscriptions of the service than users in Group C. Furthermore, we identify several significant differences in the usage of linguistic cues across the experimental groups. We finally establish significant links between the usage of linguistic cues and users' purchase behavior.


Our \textbf{main contributions} are as follows:
\begin{itemize}
  \item We provide what is, to the best of our knowledge, the first field experiment on GenAI in travel itinerary planning, examining how users respond to AI-generated itineraries in a real-world, consumer-facing context. 
  \item We demonstrate how the communication style of the AI agent (enthusiastic vs. neutral) shapes user engagement, experience, and purchase behavior.  
  \item We derive design implications for the development of GenAI systems that support complex planning tasks in customer service contexts.  
\end{itemize}


The following section provides an overview of prior work on GenAI for travel itinerary planning and AI in customer service, which motivates our research questions. After describing the field experiment and the online travel planning platform, we present the results and conclude with a discussion of implications, limitations, and directions for future research.



\section{Background}
\label{sec:rw}


\subsection{Travel Itinerary Planning with Generative AI}
\label{sec:tip}


The topic of travel itinerary planning has attracted considerable attention in prior work. 
Early research approached the task primarily as a mathematical optimization problem \citep[e.g.,][]{Bolzoni.2014,Roy.2011}. 
While these studies proposed powerful solvers and approximations, they typically relied on large structured databases containing detailed information about points of interest, cities, and regions. 
Moreover, formulating the optimization problem is a demanding task that requires technical expertise, making it inaccessible for most end users. 
Even once solved, itineraries need to be recomputed from scratch whenever users request changes, which results in long waiting times and limits the practicality of such methods in real-world customer-facing settings.


Given the challenges of traditional travel itinerary planning, researchers analyzed the usage of GenAI in itinerary planning tasks \citep[e.g.,][]{Banerjee.2025,He.2025,Xie.2023,Zhang.2024}. In fact, GenAI appears as a natural one-fits-all-solution, which should account for all of the aforementioned problems. LLMs like OpenAI's GPT models, can generate and access knowledge about countries, regions, cities, and points of interests, while supporting natural and convenient user interaction. However, \citet{Banerjee.2025} identify hallucination as a major issue, which might result in non-sensical itinerary suggestions. As a remedy, \citeauthor{Banerjee.2025} ground each user query in a knowledge basis to ensure feasibility of the proposed trips. Similarly,  \citet{He.2025} find that GenAI struggles to propose itineraries that match the user's preferences.


Furthermore, there are studies that focus on AI travel agent benchmarking and evaluation methodologies, employing meticulously curated and labeled datasets \citep{Shen.2025,Xie.2024.Preprint}. \citet{Shen.2025} introduces TripTailor, a large-scale benchmark for personalized travel planning based on real-world data from 40 Chinese cities, 500k+ POIs, and nearly 4,000 itineraries. Their experiments show that even state-of-the-art models like GPT-4o generate itineraries that rarely match human plans: only ~21\% satisfy feasibility and rationality constraints, and fewer than 10\% surpass human itineraries in personalization. The main reasons are that current GenAI agents hallucinate or confuse information, struggle with spatial–geographic reasoning (leading to inefficient routes), and fail to capture personalized user preferences—making them unsuitable for fully autonomous, high-quality travel planning at present. In a similar vein, \citet{Xie.2024.Preprint} introduces TravelPlanner, a benchmark for testing whether LLM–based agents can handle realistic travel itinerary planning. Their experiments show that even advanced models like GPT-4 achieve only a 0.6\% success rate, because current GenAI agents fail to manage multiple constraints simultaneously, often hallucinate, and struggle with long-horizon planning such as keeping track of budgets, routes, and accommodations. In short, the authors argue that GenAI is currently unsuitable for complex real-world planning tasks because it cannot reliably integrate information, respect constraints, or reason comprehensively across multi-step decision processes. 


However, there are also preprint studies by Google \citep{Udandarao.2025} and Meta \citep{Ju.2024} that demonstrate the potential feasibility of GenAI in itinerary planning based on prototypes. Roamify \citep{Udandarao.2025} is a Chrome-extension travel assistant that uses LLMs (e.g., Llama) to generate personalized itineraries based on user preferences, integrating up-to-date blog/web scraped content and recommendation systems. In user surveys and interviews, Roamify was generally preferred across age groups over traditional travel planning methods, especially when itineraries adapt dynamically to what users like. To the Globe \citep{Ju.2024} is a real-time system that takes natural language requests and translates them into symbolic constraints, which are then solved via Mixed Integer Linear Programming  to produce travel itineraries that satisfy interdependent constraints such as flights, hotels, attractions.





The study by \citet{Zhang.2024} is similar to ours. The authors analyzed the effects of chatbots expressing emotional concerns (e.g., ``The museum is crowded and not very large. I’m worried it will affect your viewing experience'') and using emoji in the context of personalized tourist attraction recommendations. The chatbot interacted with users in three steps: first, it asked about the city the user wants to visit; second, it elicited information about travel companions and budgets; and third, it proposed a tourist attraction. Their results based on three online experiments suggest that the expression of concern creates an illusion of warmth, increasing trust and satisfaction.

\subsection{Design of AI in Customer Service}
\label{sec:ai_service}


Prior research has extensively studied language in customer service when service agents are represented by humans \citep[e.g.,][]{Gremler.2020,Kranzbuhler.2020,Packard.2021,Pugh.2001} or human-AI hybrids \citep[e.g.,][]{Gnewuch.2024}. While \citet{Gremler.2020} and \citet{Kranzbuhler.2020} highlight the importance of positive emotions and rapport in customer–agent communication, \citet{Packard.2021} show that concrete rather than abstract language increases customer satisfaction. These findings are not contradictory but complementary, suggesting that effective customer service communication should ideally combine both warmth and concreteness. However, there is also evidence that, depending on the context, emotionally expressive language may also decrease the service employee's perceived competence \citep{Li.2019}. 


While AI systems are widely adopted in customer service for their efficiency and scalability \citep{Adam.2021}, much less is known about how design choices in their communication style (e.g., positive emotional vs neutral tone) shape user experience and engagement. In this regard, the review study by \citet{Diederich.2022} identifies several gaps in research on chatbots, including the need to account for user characteristics, explore adaptive design features, and investigate the long-term effects of different communication styles. They further emphasize that verbal design elements, such as emotional tone, should be studied in light of user perceptions and context. 


Several prior works investigated the design of AI in customer service \citep{Diederich.2022,Gnewuch.2017,Xie.2024}. 
Similar to human service agents, there also seems to be a trade-off between perceived warmth and competence of the AI \citep{Gray.2012,Kull.2021,Li.2019,han.2023}. \citet{Hu.2018.CHI} emphasize that tone is a crucial yet undersexplored factor in customer care chatbots. They provide empirical evidence that tones used by agent in their responses could significantly affect user experience in customer care. 
Similarly, \citet{Kull.2021} show that the tone of a chatbot’s initial message plays a crucial role in shaping brand engagement. Across three experiments, they find that a warmly phrased introduction (friendly, welcoming) reduces perceived brand–self distance and thereby fosters stronger consumer engagement compared to competent or neutral messages. This effect is particularly pronounced when consumers already have a neutral or positive relationship with the brand. 
\citet{Li.2019} examine how the use of emoticons by service employees shapes user perceptions in online customer–service interactions. Through controlled experiments and a field study, they find that emoticons increase perceived warmth but decrease perceived competence. The work highlights how subtle paralinguistic cues in text-based interfaces can affect satisfaction and behavioral responses, underscoring the need for context-sensitive design of social signals in service technologies. 
\citet{han.2023} hypothesize that customers have prior experiences regarding AI's capability of feeling and expressing emotions. Given that AI is substantially different from humans because it is not perceived to possess experience (i.e., the capacity to feel emotions), humans should be more likely to expect neutral instead of emotional language \citep{Gray.2012}. Although AI is increasingly becoming more sophisticated in recognizing human emotions \citep{Mello.2013,Wang.2022}, it remains unclear how humans respond when communicating with emotionally-charged AI in different contexts. 


Setting the chatbot's language to meet the user's expectation for particular contexts is considered a high design priority \citep{Chaves.2020}. \citet{Chaves.2020} proposed a framework for register-adaptive utterance that align with users' expectations, which could enhance perceived credibility, appropriateness, and social role attribution. Collectively, these findings suggest that tailoring conversational tone to users' contextual expectations is a key determinant of chatbot effectiveness as an assistant. So far, the investigated contexts include, among others, tourist attraction recommendation \citep{Zhang.2024}, psychotherapy \citep{Song.2025}, customer care \citep{Hu.2018.CHI}, movie recommendations \citep{Lee.2017}, and feedback collection \citep{Adam.2021}.




\subsection{Research Questions}
\label{sec:rq}

In summary, prior research has identified major challenges to using GenAI for travel itinerary planning \citep{He.2025,Shen.2025,Xie.2024}, but prototypes suggest potential feasibility \citep{Ju.2024,Udandarao.2025}. Studies on AI in customer service have produced mixed results regarding the role of language \citep{Gray.2012,Kull.2021,han.2023}.  

Following \citet{Chaves.2020}, we argue that users’ expectations about AI language are a critical design choice that depends on context. Travel itinerary planning involves high cognitive effort (optimizing time and budget) but also emotional aspects (excitement and fear of missing out).  

We therefore examine whether GenAI should adopt an \textit{enthusiastic} versus a \textit{neutral} tone when communicating with users. In a field experiment with paying participants, we analyze how tone affects (1) linguistic features of user prompts, (2) user experience, and (3) purchase behavior. Our research questions are:  

\begin{itemize}[leftmargin=1.5em]
  \item \textbf{RQ1:} How does the communication tone of GenAI (enthusiastic vs. neutral) influence user engagement and purchase behavior in travel itinerary planning? 
  \item \textbf{RQ2:} How does GenAI’s tone shape the linguistic characteristics of users’ input prompts? 
  \item \textbf{RQ3:} How are linguistic cues in user prompts related to subsequent purchase decisions?
\end{itemize}

\section{Method}
\label{sec:method}

\begin{figure}[htp]
    \centering
    \includegraphics[trim=0cm 5.5cm 0cm 2cm, clip, width=\textwidth]{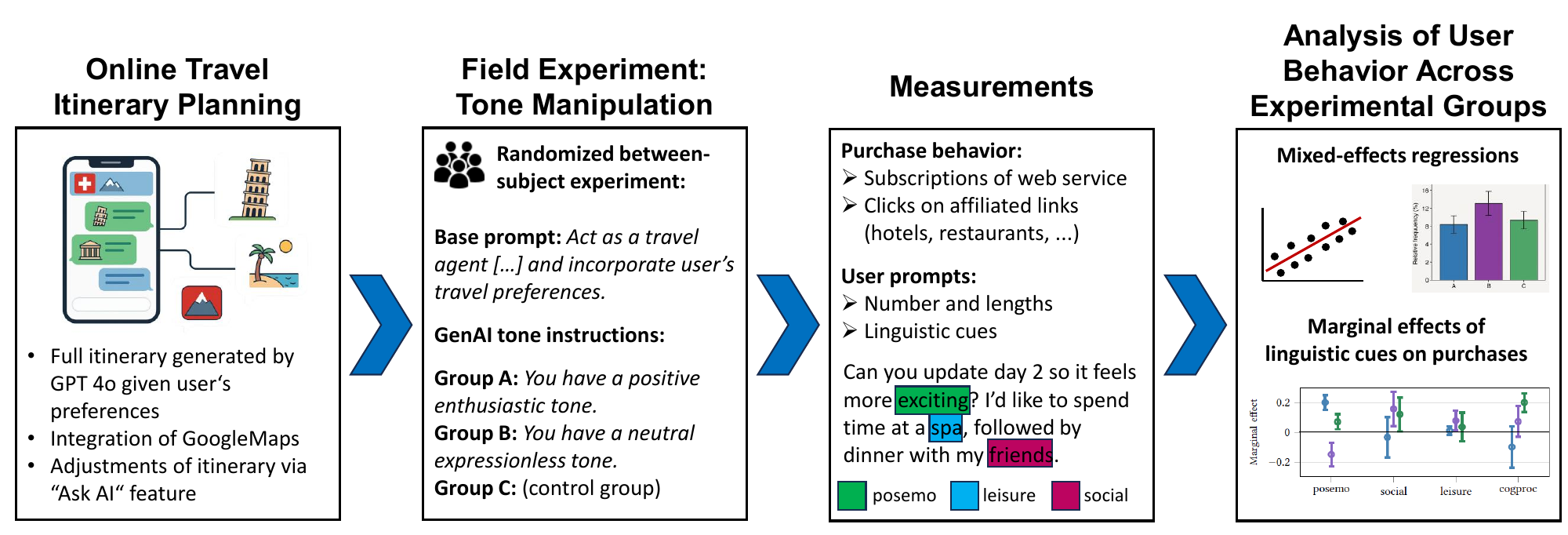}
    \caption{Research framework.}
    \label{fig:framework}
\end{figure}

Our research framework is illustrated in \Cref{fig:framework}. We performed a randomized field experiment in an existing webservice about online travel itinerary planning. Specifically, we manipulated the tone instructions added to the base prompt, which guides the AI to adjusting the itinerary based on the user's preferences. We included three experimental groups, namely, Group~(A) positive enthuaiastic tone, Group~(B) neutral expressionless, and Group~(C) with no tone manipulation as control group. We measure the length and number user input prompts, as well as linguistic cues used by users. In addition, we store data about users' purchase behavior in terms of subscriptions of the webservice and clicks on affiliated links. Finally, we estimate mixed-effects regression models to analyze differences in user engagement, purchase behavior, and how users' usage of linguistic cues is linked to subscriptions and clicks on affiliated links.


\subsection{Platform and Procedure}



We conducted a randomized field experiment to evaluate user interactions with a Generative AI–powered chat feature embedded in the \emph{Tern.app} travel planning platform. 
Using OpenAI GPT models, Tern generates highly personalized and customized travel plans. The user interface and the chatbot allow easy editing and collaboration, allowing users to easily adjust their itineraries and share them with others.

\begin{figure}[htp]
    \centering
    \includegraphics[width=0.7\textwidth]{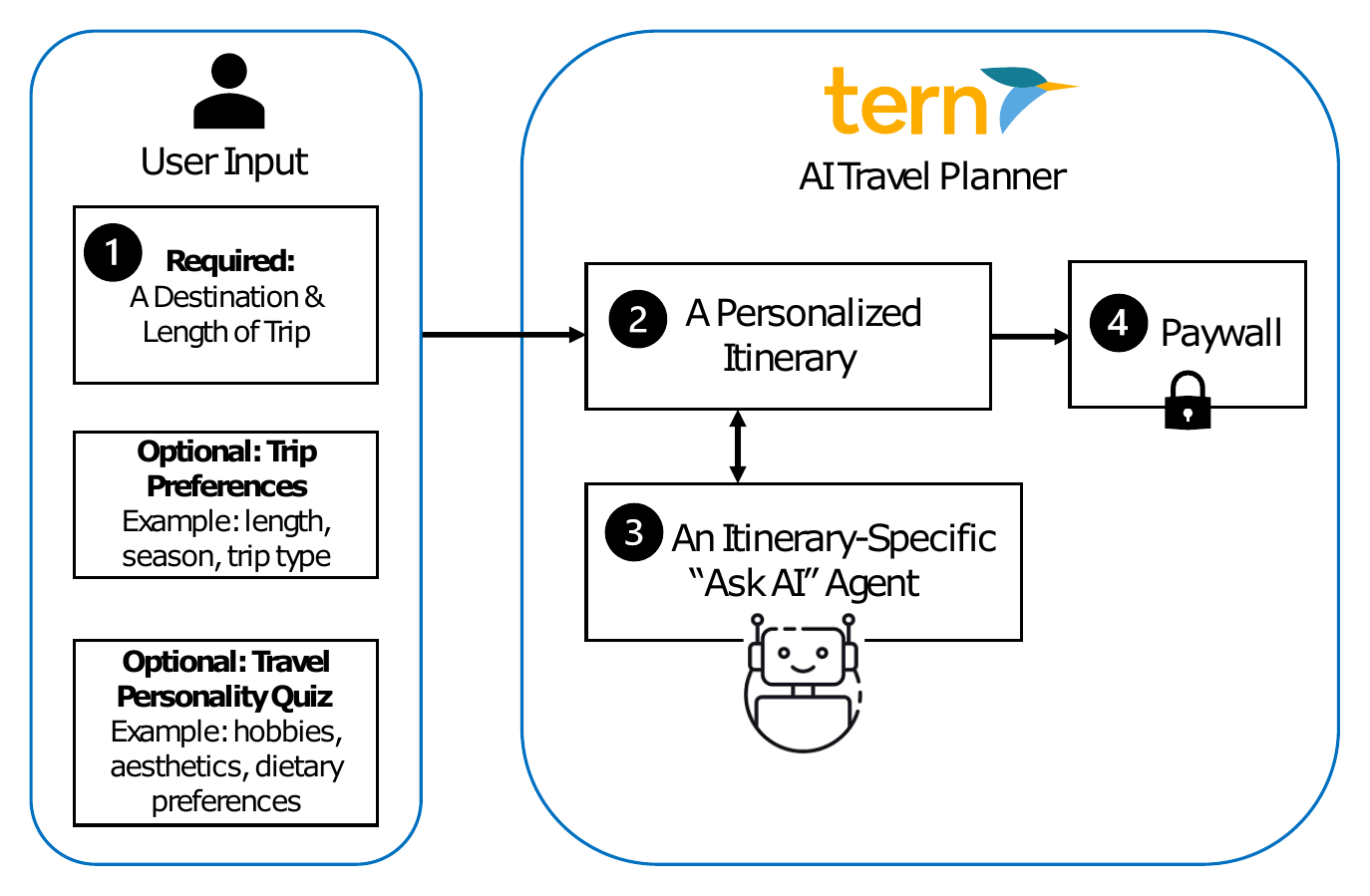}
    \caption{Itinerary planning procedure on tern.}
    \label{fig:platform}
\end{figure}

The planning procedure on the platform is illustrated in \Cref{fig:platform}. 
Upon visiting the Tern platform \href{https://tern.app}{tern.app}, the user inputs a destination. There is a dropdown menu that let the user input trip length, season, trip type, and budget. All first-time users are presented with an option to take a Travel Personality Quiz that gives recommendations on destinations based on answers about hobbies, aesthetics and dietary preferences. The quiz’s purpose is to enhance personalization of the itinerary and any future itineraries. Tern then generates a personalized itinerary. The user may choose to engage with ”Ask  AI” to get additional information about the destination or edit their itinerary. After two days of access to the itinerary, the user is asked to subscribed.



Before accessing their personalized itinerary, users are invited to complete a five-question travel personality quiz. Participation in the quiz is optional. Subsequently, users are prompted to create a Tern account using their email address, phone number, or Google account. First-time users who complete the quiz are shown their personalized results, which include descriptive insights and suggested travel destinations. The itinerary is then generated by GPT 4o. To ensure the validity of generated content, itineraries are cross-validated against the Google Maps API. The API not only confirms the existence of the suggested points of interest and sights but also provides corresponding images. This integration effectively mitigates hallucinations common in generative models and ensures that the resulting itineraries remain coherent and contextually meaningful.



\begin{figure}[htp]
    \centering
    \includegraphics[width=\textwidth]{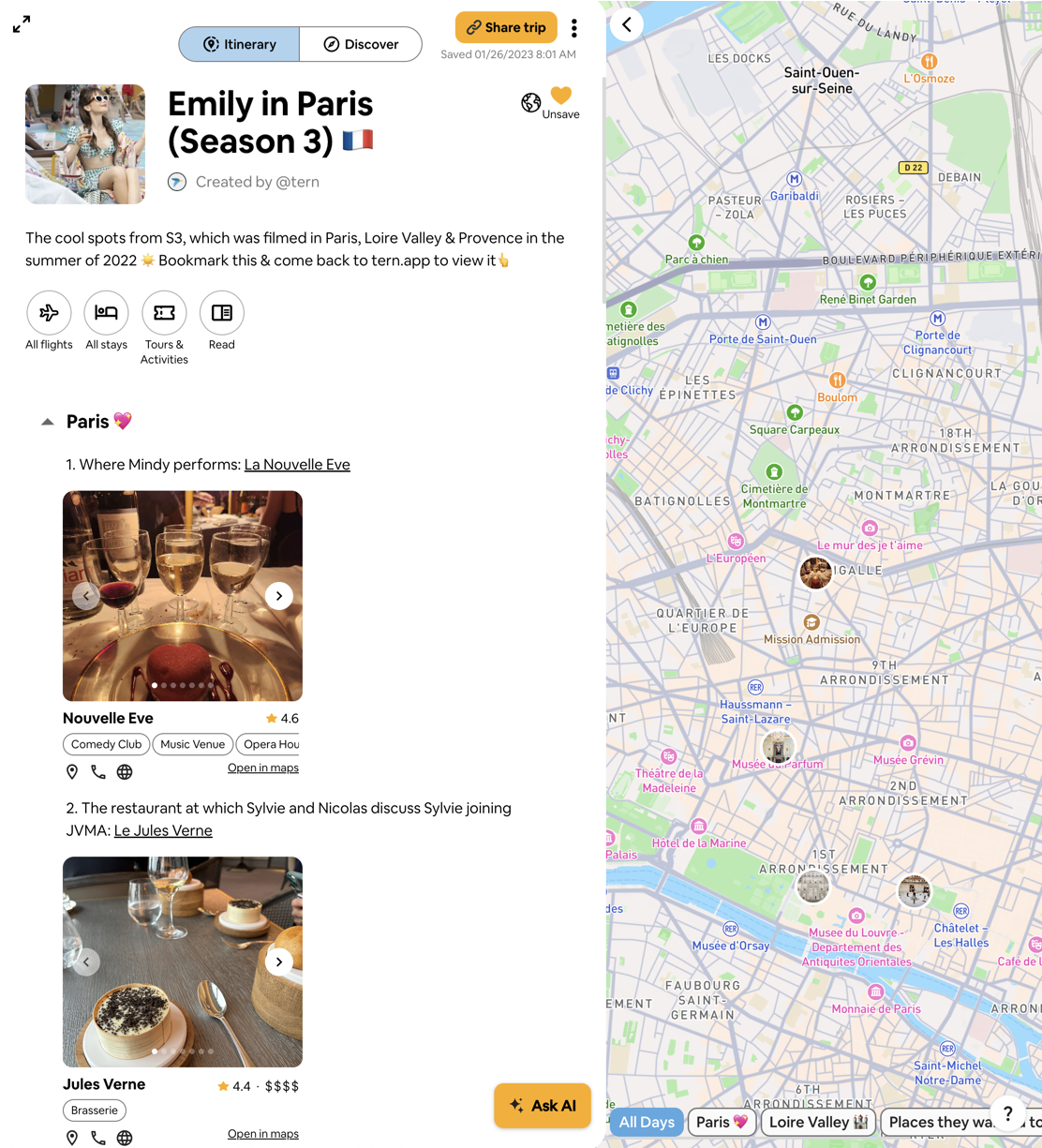}
    \caption{User interface of itinerary overview.}
    \label{fig:tern}
\end{figure}

\begin{figure}[htp]
\centering
    \includegraphics[width=0.75\linewidth]{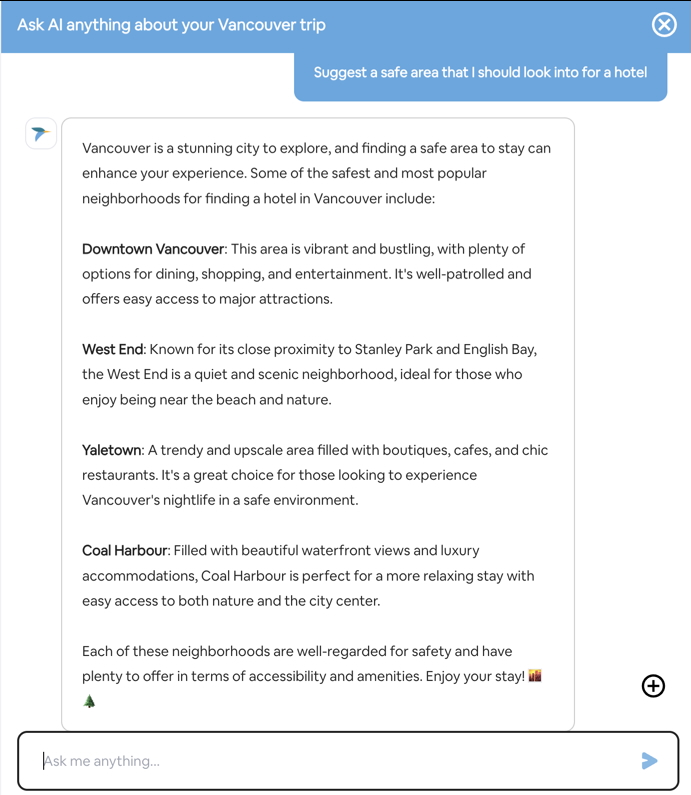}
    \caption{User interface of the ``Ask AI'' feature. \label{fig:gui}}
\end{figure}

Once the itinerary is generated, users are presented with a day-by-day schedule of recommended activities and dining options. The user interface shown in \Cref{fig:tern} illustrates the daily itinerary schedule. A dynamic map is displayed on the right-hand side, populated with pin drops marking points of interest. An ``Ask AI'' button is anchored to the bottom of the interface, enabling users to request itinerary modifications or obtain additional information about their destination through AI-assisted interaction. The images of restaurants, sights, and other POIs are presented as affiliated links. 
The user interface of the ``Ask AI'' feature is shown in \Cref{fig:gui}.

Tern has been adopted by a global user base, with thousands of trips planned across more than 120 countries. A significant portion of the users are based in the United States, comprising approximately 42\% of the total. Users who signed up prior to October 2024 have access to the platform free of charge; however, a paywall has since been introduced that offers two pricing subscription options. Users can choose between purchasing an unlimited access to one itinerary with no expiration date (\$5 per itinerary), or an annual subscription (\$29 per year) for unrestricted access to create unconstrained amount of itineraries. Subscription access includes the ability to edit, save, and share itineraries with unlimited collaborators. Both options include access to the "Ask AI" travel agent chatbot. The service is available to users worldwide.

\subsection{Participants}


The experiment commenced on November 3, 2024, and concluded on April 14, 2025, spanning approximately 4.5 months of data collection. During this period, the platform attracted 5,212 unique visitors. Of these users, 1,774 (34\%) proceeded to create a user profile and 1,617 (31\% of total visitors) successfully generated at least one travel itinerary. There were 670 users that clicked on the ``Ask AI''  button, but not all proceeded to chat.


A total of 282 users chatted with the AI. The majority of these users were based in the United States (94 users, 34\%), followed by users from Europe (43), India (20), the United Kingdom (12), Australia (11), Canada (9), Mexico (2), Brazil (2), and Singapore (1). For 58 users, the country was not identified. In terms of age distribution, 102 users (36\%) were between the ages of 26--40, 49 users (17\%) were aged 41 and above, 43 users (15\%) were between 18--25, and 4 users (1\%) were under the age of 18. Age data was unavailable for 49 users (18\%). No gender information is available as it was not asked by the user interface. Regarding the devices, 124 used laptops or desktop computers (62 Windows, 58 Mac OS X, 4 Linux), 99 users had mobile devices (35 Android, 64 iOS) and 59 users had unknown devices. We collected 1301 input queries from the users who chatted with the AI Agent.


We ensured anonymity of the data by removing any personal information. For this purpose, we implemented Personally Identifiable Information (PII) filtering using the presidio-analyzer Python library, which is consistent with the methodology proposed by \citet{Zhu.2025} for post-processing to detect and remove PII.  Unintentionally leaked PII was subsequently filtered manually.

\subsection{Manipulation: Tone Instructions in System Prompt}

The ``Ask AI'' feature is powered by OpenAI's GPT-4o model and guided by a system-level prompt that instructs it to emulate the role of a knowledgeable and helpful travel agent, providing personalized recommendations, itinerary adjustments, and destination-specific insights in response to user queries.

All users were randomly assigned to one of three experimental conditions, each defined by a variation in the tone of the AI assistant's prompt. 
\begin{itemize}
    \item \textbf{Group A (Positive enthusiastic)}: \emph{You have a positive enthusiastic tone. Your persona is a charming hotel concierge. Use emojis when appropriate.}
    \item \textbf{Group B (Neutral expressionless)}: \emph{You have a neutral expressionless tone.}
    \item \textbf{Group C (Control)}: \emph{(No tone-instruction augmentation to the base prompt.)}
\end{itemize}

The underlying model and the base prompt content remained constant across conditions, with only the tone instructions augmented. The base prompt is the startup's business secret, but in essence, it consists of the following instructions:

\begin{list}{}{\leftmargin=2em \rightmargin=2em}
\item
\emph{Act as a travel agent [...] and incorporate user's travel preferences.}
\end{list}

All other system parameters, including model configuration and interface design, were held constant. Of the 282 users that chatted with the AI, 86 were randomly assigned to Group A, 105 to Group B, and 91 to Group C.

\subsection{Example Conversations}

We provide examplary conversations for users in each group. Additional example conversations are provided in Appendix A.

\subsubsection*{Group A}

\begin{userbubble}\footnotesize
\userlabel{A}: \emph{``Update (Tue Aug 19) Day 5: Dingle Peninsula with different things to do around the cliffs of moher, burren, and limerick, ending in Killarney''}
\end{userbubble}

\begin{aibubble}\footnotesize
\ailabel: Here's your updated Day 5 itinerary with some exciting new experiences around the Cliffs of Moher, the Burren, and Limerick, ending in the charming town of Killarney:\\
\textbf{(Tue Aug 19) Day 5: Cliffs of Moher, Burren, and Killarney Journey}
\begin{itemize}\itemsep0.2em
  \item \textbf{Morning}: Start your day at the Cliffs of Moher, exploring the scenic cliffside paths with guided tours for the best views.
  \item \textbf{Mid-Morning}: Venture into Burren National Park for a guided walk exploring its unique karst landscape and fascinating flora.
  \item \textbf{Afternoon}: Head to Limerick and enjoy lunch at a local spot. Visit King John's Castle to learn about the city's medieval past.
  \item \textbf{Late Afternoon}: Travel to Killarney, a picturesque drive through Ireland's countryside.
  \item \textbf{Evening}: Arrive in Killarney, check into your accommodation, and enjoy dinner at a cozy local restaurant.
\end{itemize}
This update includes breathtaking natural wonders, historical insights, and a delightful drive, perfect for an explorer like you who loves diving into the heart of Ireland! \emoji{herb} \emoji{castle} \emoji{automobile}
\end{aibubble}

\begin{userbubble}\footnotesize
\userlabel{A}: \emph{``What should we do on the way to Kilkenny from Dublin''}
\end{userbubble}

\begin{aibubble}\footnotesize
\ailabel: You're in for a delightful journey from Dublin to Kilkenny! \emoji{automobile} As an adventurous and curious traveler who enjoys hiking and outdoor adventures, here are a few stops that would pique your interests:
\begin{itemize}\itemsep0.2em
\item \textbf{Glendalough}: Nestled in the Wicklow Mountains, this serene and tranquil location offers beautiful hiking trails surrounding ancient monastic sites. The trails vary in length and difficulty, so you can choose what suits you best.
\item \textbf{The Wicklow Mountains National Park}: Explore some breathtaking landscapes with numerous walking routes. Don't miss the scenic Sally Gap drive, which offers picturesque views ideal for nature lovers like yourself.
\item \textbf{Powerscourt House and Gardens}: This stop is perfect for appreciating luxury and elegance in nature. Stroll through the beautifully manicured gardens, enjoy the waterfall, or have a cozy lunch in their café.
\item \textbf{Mount Usher Gardens}: These gardens offer a relaxed and natural vibe, perfect for a peaceful walk amidst exotic plants and the River Vartry's gentle flow.
\item \textbf{Jerpoint Abbey} - As you get closer to Kilkenny, make a quick stop to explore these historical ruins. It's a great place to delve into local history and enjoy some tranquil moments before reaching your destination.
\end{itemize}
I hope you have a fantastic trip filled with adventure and tranquility! \emoji{herb}
\end{aibubble}

\subsubsection*{Group B}
\begin{userbubble}\footnotesize
\userlabel{B}: \emph{``What are the best flights from BDL to Lisbon?''}
\end{userbubble}
\begin{aibubble}\footnotesize
\ailabel: For flights from \textbf{Bradley International Airport (BDL)} to Lisbon, it's best to consider airlines that offer convenient routes with minimal stopovers. Here are some options:
\begin{itemize}
\item \textbf{Delta Airlines}: Offers flights with one stop, typically connecting through major hubs like New York City (JFK).
\item \textbf{United Airlines}: Connects through hubs such as Washington D.C. (IAD) and New York City (EWR).
\item \textbf{TAP Air Portugal}: Though direct flights may not be available from BDL, TAP Air Portugal often provides routes from nearby major airports on the East Coast. 
\end{itemize}
For the most accurate and current options, check with airlines directly or use flight comparison websites to find the best routing and pricing for your specific travel dates.
\end{aibubble}

\subsubsection*{Group C}

\begin{userbubble}\footnotesize
\userlabel{C}: \emph{``Non-alcoholic fondue in Basel, Switzerland''}
\end{userbubble}
\begin{aibubble}\footnotesize
\ailabel: I found some information for your itinerary in Basel, Switzerland, and also some insights into your preferences from your quiz responses. Based on the quiz you took, you seem to enjoy adventures, nature, art, quirky aesthetics, and a variety of food styles. Here's an updated itinerary suggestion for your trip:
\textbf{Art, Fondue, and Creativity (Day 3)}
\begin{itemize}
    \item \textbf{Morning}: Visit the \textbf{Fondation Beyeler}, a renowned art museum located just outside Basel, featuring a stunning collection of modern and contemporary art.
    \item \textbf{Afternoon}: Explore the quirky and eclectic \textbf{Vitra Campus} in nearby Weil am Rhein, Germany. The campus is home to unique architecture and design exhibitions.
    \item \textbf{Evening}: Experience a non-alcoholic fondue dinner at a local restaurant such as \textbf{Walliser Kanne}, which offers a cozy, traditional ambiance perfect for enjoying this Swiss classic, ensuring options are in line with your preferences.
\end{itemize}
This itinerary combines your interests in art, nature, culinary experiences, and quirky aesthetics.
\end{aibubble}


\subsection{Measurements}

We store all user input prompts to the ``Ask AI'' feature. The number and length of the prompts serve as a measure of user engagement. To capture aspects of user experience, we extract linguistic cues from the prompts, which reflect users’ emotional, cognitive, and social orientation during interaction \citep{Jang.2017}. Finally, we count the number of annual subscriptions and clicks on affiliate links as behavioral measures of purchases.

\subsection{Ethical Considerations}

The field experiment analyzed anonymized behavioral data from users of an itinerary planning platform. 
Participants were not explicitly informed that they were part of an experiment, as the interventions were 
implemented as part of the regular functionality of the platform. 
All data were collected in anonymized form, contained no personally identifiable information, 
and were analyzed in aggregate. 
According to institutional and legal regulations, no formal ethics approval was required for this field study. 
At the same time, we plan to replicate the study under controlled laboratory conditions including neurophysiological measurements of cognitive and affective processes. 
For these settings, our research group has obtained ethics approval from the University of (anonymized), Approval No.~24-1533-S2, 
which  underscores our commitment to conducting research in line with established ethical standards.

\section{Results}
\label{sec:results}

\subsection{User Engagement}

We start by analyzing user engagement across the experimental groups regarding the length and number of input prompts. Our data contains 282 users that chatted with the AI. Out of these, 86 were randomly assigned to Group A, 105 to Group B, and 91 to Group C. The number of input prompts is 404 in Group A, 506 in Group B, and 391 in Group C.

\subsubsection*{Length of Input Prompts}

We use mixed-effects regression models with random-effects at the user-level (equivalent to one-way ANOVA with repeated measurements) to analyze the length and number of input prompts.\footnote{All analyses were performed in Python v3.12. We used the \texttt{mixedlm} function with Wald z-tests for fixed effects from \texttt{statsmodels}.} Thereby, we model the user-specific repeated measurements across input prompts while allowing sufficient power to identify significant differences \citep{Baayen.2008}. Different from fixed effects, including a random effect still allows us to analyze the between-group variation \citep{Townsend.2013} given by the AI's tone instructions. Accordingly, we estimate a mixed-effects OLS regression model with random intercept $\mu_U$ at the user level to explain the length of input prompts
\begin{align}
    \text{Prompt length} &= \alpha + \beta_{B}\,1_{\text{Group}=B} 
     + \beta_{C}\,1_{\text{Group}=C} + \mu_U + \varepsilon.\label{eq:model_ols_length}
\end{align}
The reference of the categorical Group variable is given by Group A.

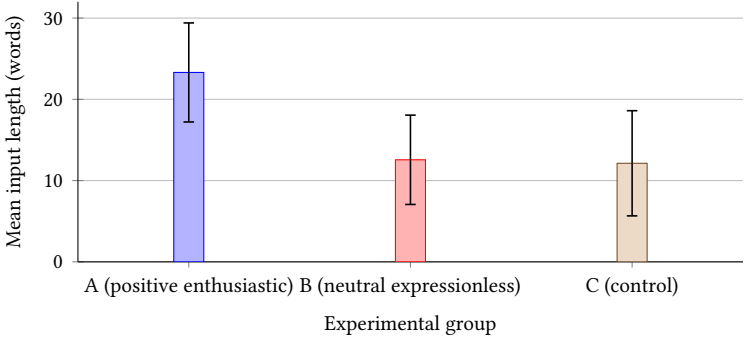
\begin{figure}[htp]
    \centering
    \vspace{0.3cm}
\begin{tikzpicture}
\pgfplotsset{
  error bars/.cd,
  y dir=both, y explicit,
  error bar style={draw=black, line width=0.6pt},
  error mark options={draw=black, line width=0.6pt},
  error mark=|
}
\begin{axis}[
  width=0.75\linewidth,
  height=5cm,
  ybar, bar shift=0pt,
  bar shift auto=false,
  bar width=0.4cm,                 
  ymajorgrids,
  tick pos=lower, tick align=outside,
  axis lines=left, axis line style={-},
  ymin=0, ymax=32,
  xtick={1,2,3},                 
  xticklabels={
    A (positive enthusiastic),
    B (neutral expressionless),
    C (control)
  },
  x tick label style={font=\footnotesize},
  y tick label style={font=\footnotesize},
  xlabel={Experimental group},
  xlabel style={font=\footnotesize},
  ylabel={Mean input length (words)},
  ylabel style={font=\footnotesize},
  enlarge x limits=0.25          
]

 \addplot+[bar shift=0pt] coordinates {(1,23.31) +- (0,6.094)};
  \addplot+[bar shift=0pt] coordinates {(2,12.56) +- (0,5.497)};
  \addplot+[bar shift=0pt] coordinates {(3,12.13) +- (0,6.472)};

\end{axis}
\end{tikzpicture}
    \caption{Input prompt length (in words) across groups estimated by mixed-effects regression model. The error bars denote the 95\% confidence intervals.}
    \label{fig:length}
\end{figure}

The estimated marginal mean lengths in words with 95\% confidence intervals are shown in \Cref{fig:length}. Users in Groups B and C produced significantly shorter prompts than those in Group A ($\beta_B = -10.75, 95\%\ \text{CI}\ [-18.97,-2.54], p = .01;\, \beta_C=-11.18, 95\%\ \text{CI}\ [-19.98, -2.39], p=.013$).


\subsubsection*{Number of Input Prompts per User}

\begin{figure}[htp]
    \centering
    \vspace{0.3cm}
\begin{subfigure}[b]{0.49\textwidth} 
\begin{tikzpicture}
\begin{axis}[
  ymajorgrids, width=\linewidth, height=5cm,
  ylabel={Number of input prompts},
  xtick={1,2,3}, 
  ymax=150,
   tick label style={font=\footnotesize},
  label style={font=\footnotesize},
  xlabel={Experimental group},
   xticklabels={A, B , C}, 
   boxplot/draw direction=y
]
\addplot+[
  boxplot,
  boxplot/draw position=1
] table[y=n_inputs, col sep=comma] {user_counts_A.csv};
\addplot+[
  boxplot,
  boxplot/draw position=2
] table[y=n_inputs, col sep=comma] {user_counts_B.csv};
\addplot+[
  boxplot,
  boxplot/draw position=3
] table[y=n_inputs, col sep=comma] {user_counts_C.csv};
\end{axis}
\end{tikzpicture}
\end{subfigure}
\begin{subfigure}[b]{0.49\textwidth} 
\begin{tikzpicture}
\begin{axis}[
  ymajorgrids, width=\linewidth, height=5cm,
  ylabel={Number of input prompts},
  xtick={1,2,3}, ymax=10.5,
   tick label style={font=\footnotesize},
  label style={font=\footnotesize},
   xlabel={Experimental group},
  xticklabels={A, B , C}, 
  boxplot/draw direction=y
]
\addplot+[
  boxplot, 
  boxplot/draw position=1
] table[y=n_inputs, col sep=comma] {user_counts_A.csv};
\addplot+[
  boxplot, 
  boxplot/draw position=2
] table[y=n_inputs, col sep=comma] {user_counts_B.csv};
\addplot+[
  boxplot, 
  boxplot/draw position=3
] table[y=n_inputs, col sep=comma] {user_counts_C.csv};
\end{axis}
\end{tikzpicture}
\end{subfigure}

\caption{Boxplots showing number of prompts per user across experimental groups.}
\label{fig:num_prompts}
\end{figure}
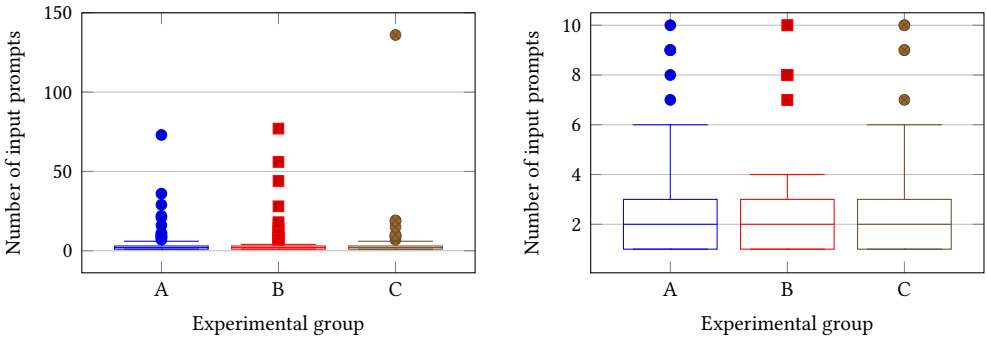

Next, we analyze the number of input prompts per user across experimental groups. For this purpose, we estimate the following mixed-effects regression model
\begin{align}
\text{Number of prompts}_U &= \alpha  + \beta_{B}\,1_{\text{Group}=B} 
     + \beta_{C}\,1_{\text{Group}=C}+ \mu_U + \varepsilon.\label{eq:model_ols_num} 
\end{align}
We find no significant differences across the experimental groups as neither $\beta_B$ nor $\beta_C$ are statistically significant. In fact, the distributions of the number of inputs per user are very similar as shown in \Cref{fig:num_prompts}. The figure shows the boxplots of the number of input prompts per user for all groups over the entire domain (left) and zoomed in to the range 1--10 (right). The median number of inputs was 2 and the third quartile 3 per user across all groups, with outliers up to 136 prompts.

\subsection{Subscriptions and Clicks on Affiliated Links}

We proceeed by analyzing purchase behavior in terms of subscriptions to the itinerary planning service and clicks on affiliated links. The data is aggregated at the user level with binary outcome variables: a user is labeled positive if they subscribed or if they clicked on at least one affiliated link. The resulting dataset consists of 86 observations for Group A (13 subscribers, 17 users with at least one click), 105 for Group B (15 subscribers, 21 users with at least one click), and 91 for Group C (5 subscribers, 18 users with at least one click). Since outcomes are defined at the user level, we estimate standard logistic regression models of the form
\begin{equation}
\log \left(\frac{\Pr(Y = 1)}{\Pr(Y = 0)} \right)
= \alpha + \beta_{B}\,1_{\text{Group}=B}
+ \beta_{C}\,1_{\text{Group}=C},
\label{eq:model_logreg}
\end{equation}
where $Y \in \lbrace \text{Subscription}, \text{Click on Affiliated Link} \rbrace$ and Group A serves as the reference.\footnote{We used the \texttt{logit} function from \texttt{statsmodels}.}





\begin{figure}[htp]
\begin{tikzpicture}
\pgfplotsset{
  error bars/.cd,
    y dir=both, y explicit,
    error bar style={draw=black, line width=1.2pt},   
    error mark options={draw=black, line width=1.2pt},
    error mark=|,  
}
\begin{axis}[
  ybar,
  ymin=0, ymax=0.35,
  ylabel={Subscription / click rate},
  symbolic x coords={A,B,C}, 
  xticklabels={A (positive ethusiastic), B (neutral expressionless), C (control)},
  xtick=data,
  y tick label style={font=\footnotesize},
    x tick label style={font=\footnotesize},
    xlabel={Experimental group},
    xlabel style={font=\footnotesize},
    ylabel style={font=\footnotesize},
  bar width=12pt,
  ymajorgrids,
  axis lines=left, axis line style={-},
  width=0.8\linewidth, height=5cm,
  enlarge x limits=0.25,
  nodes near coords,
  nodes near coords align={vertical},
  every node near coord/.append style={font=\footnotesize, yshift=2pt},
  nodes near coords style={/pgf/number format/.cd,fixed,precision=2}, 
]

\addplot+[
  draw=black, fill=niceblue]
table[
  col sep=comma,
  x=groupType,
  y=rate,
  y error plus expr=\thisrow{ci_up}-\thisrow{rate},
  y error minus expr=\thisrow{rate}-\thisrow{ci_low},
  point meta=explicit symbolic,
]{sub_rates.csv}; \label{pgf:subs}

\addplot+[
  draw=black, fill=nicered]
table[
  col sep=comma,
  x=groupType,
  y=rate,
  y error plus expr=\thisrow{ci_up}-\thisrow{rate},
  y error minus expr=\thisrow{rate}-\thisrow{ci_low},
  point meta=explicit symbolic,
]{click_rates.csv}; \label{pgf:links}

\pgfplotsset{every node near coord/.append style={/pgfplots/point meta=explicit symbolic}}

\end{axis}
\end{tikzpicture}

\vspace{0.2cm}

\begin{tikzpicture}
    \matrix[matrix of nodes, draw, 
            nodes={anchor=west, font=\footnotesize},
            column sep=0.5em,
            row sep=0em,
            inner sep=2pt, 
            align=left,
            ampersand replacement=\&] at (current bounding box.south) {

   \ref{pgf:subs} Subscriptions  \& \ref{pgf:links} Clicks on Affiliated Links\\
    };
  \end{tikzpicture}

    \vspace{0.3cm}
    \abovecaptionskip=0pt
    \caption{Subscriptions and clicks on affiliated links with 95\% confidence intervals across experimental groups.}
    \label{fig:buy}
\end{figure}
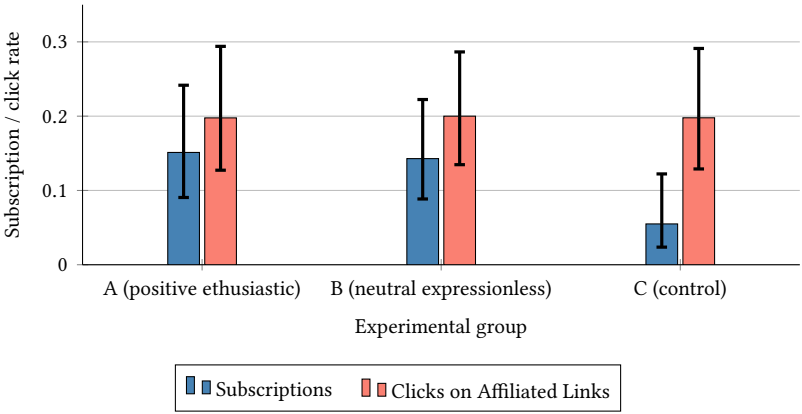

The predicted probabilities of subscriptions and clicks on affiliated links with 95\% confidence intervals are shown in \Cref{fig:buy}. For subscriptions, users in Groups A and B were significantly more likely to subscribe compared to users in Group C (predicted probability: Group A = 15.1\%, Group B = 14.2\%, Group C = 5.5\%). The estimated coefficients of the logistic regression are $\beta_B = -0.067$ ($95\%\ \text{CI}\ [-0.87, 0.74], p = .024$) and $\beta_C = -1.12$ ($95\%\ \text{CI}\ [-2.20, -0.04], p = .042$). However, there are no significant differences for clicks on affiliated links.

\subsection{Linguistic Cues in Input Prompts}

We also analyze perform an explorative analysis to understand how the user input prompts differ across experimental groups. 
Here, we follow prior research \citep[e.g.,][]{Fang.2023,Jang.2017,Wang.2024} in calculating 
linguistic cues with ``Linguistic Inquiry and Word Count'' \citep[LIWC,][]{Pennebaker.2015}. 
This software is based on a dictionary that contains more than 4,500 words, each of which is classified 
into one or more categories of linguistic cues. Such cues can be interpreted as proxies for 
\textit{aspects of user experience}, as prior work has shown that the way users express themselves 
linguistically reflects their cognitive and emotional states \citep{Jang.2017}. Linguistic cues are then calculated as 
the fraction of words linked to a particular category (e.g., \highlightcue{positive emotion words, leisure, time}). 



Given that LIWC provides more than 90 linguistic categories, we restrict our analysis to a subset that is both frequent and informative. Specifically, we consider only those cues that occurred with a relative frequency of at least 1\% in at least one group when aggregated at the user level, following prior work on linguistic cue filtering \citep{Lutz.2024}. This selection ensures that our evaluation focuses on cues that allow for meaningful interpretation while avoiding noise from extremely rare categories. Following this approach, we obtain a total of 28 linguistic cues as shown in \Cref{tbl:cues_example}. The table provides the particular LIWC category and examples from \citet{Pennebaker.2015}.


\begin{table}[htp]
\centering
\caption{Overview of selected linguistic cues. \label{tbl:cues_example}}
\footnotesize
\setlength{\defaultaddspace}{1.5pt}
\begin{tabular}{llll} 
\toprule

\textbf{Linguistic Cue} & \textbf{Category} & {\textbf{Examples}} & {\textbf{Category Words}}  \\

\midrule

All Punctuation & \texttt{AllPunc} & -- & -- \\ \addlinespace
Dashes & \texttt{Dash} & -- & -- \\ \addlinespace
Periods & \texttt{Period}  & -- & -- \\ \addlinespace
Question marks & \texttt{QMark}  & -- & --  \\ \addlinespace
Other punctuation & \texttt{OtherP} & -- & -- \\ \addlinespace
Achievement & \texttt{achieve} & \emph{win, success, better} & 213 \\ \addlinespace
Adjectives & \texttt{adj} & \emph{free, happy, long} & 764 \\ \addlinespace
Affective processes & \texttt{affect} & \emph{happy, cried} & 1393  \\ \addlinespace 
Assent & \texttt{assent} & \emph{agree, OK, yes} & 36 \\ \addlinespace
Biological processes & \texttt{bio} &  \emph{eat, blood, pain} & 748 \\ \addlinespace
Cognitive processes & \texttt{cogproc} & \emph{cause, know, ought} & 797 \\ \addlinespace
Drives (motivational concerns) & \texttt{drives} & \emph{ally, win, risk} & 1103 \\ \addlinespace
Function & \texttt{funct} & \emph{it, to, no, very} & 491 \\ \addlinespace
Informal language & \texttt{informal} & \emph{damn, thx, lol} & 380 \\ \addlinespace
Ingestion & \texttt{ingest} & \emph{dish, eat, pizza} & 184 \\ \addlinespace
Interrogatives & \texttt{interrog} & \emph{how, when, what} & 48 \\ \addlinespace 
Leisure & \texttt{leisure} & \emph{cook, chat, movie} & 296 \\ \addlinespace
Motion & \texttt{motion} & \emph{arrive, car, go} & 325 \\ \addlinespace 
Netspeak & \texttt{netspeak} & \emph{btw, lol, thx} & 209 \\ \addlinespace
Numbers & \texttt{number} & \emph{one, seven} & 36 \\ \addlinespace 
Perception & \texttt{percept} & \emph{look, heard, feeling} & 436 \\ \addlinespace 
Positive emotion & \texttt{posemo} & \emph{love, nice, sweet} & 620 \\ 
Prepositions & \texttt{prep} & \emph{to, with, above} & 74 \\ \addlinespace
Relativity & \texttt{relativ} & \emph{area, bend, exit} & 974 \\ \addlinespace
Risk & \texttt{risk} & \emph{danger, doubt} & 103 \\ \addlinespace
Social processes & \texttt{social} & \emph{mate, friends, daughter} & 756\\ \addlinespace
Space & \texttt{space} & \emph{down, in, thin} & 360 \\ \addlinespace
Time & \texttt{time} & \emph{end, until, season} & 310 \\ \addlinespace

\bottomrule
\multicolumn{4}{l}{Note: All information is taken from the LIWC 2015 manual  \citep{Pennebaker.2015}.} \\
\end{tabular}

\end{table}

\subsubsection*{Usage Across Experimental Groups}

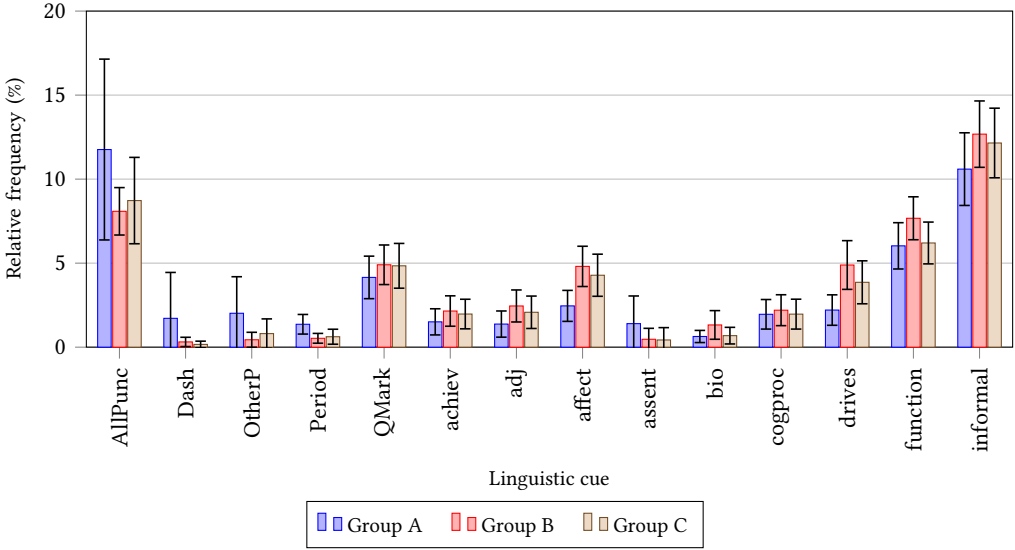
\begin{figure}[htp]
    \centering
\begin{tikzpicture}
\pgfplotsset{
  error bars/.cd,
    y dir=both, y explicit,
    error bar style={draw=black, line width=0.6pt},   
    error mark options={draw=black, line width=0.6pt},
    error mark=|,  
}

\begin{axis}[
    width=\linewidth,
    height=6cm, ybar, 
    ymajorgrids, bar width=0.18cm,
    tick pos=lower,
    tick align=outside,
    xmin=0.5, xmax=14.5,
    ymin=0, ymax=20,
    xtick={1,2,3,4,5,6,7,8,9,10,11,12,13,14},
xticklabels={AllPunc, Dash, OtherP, Period, QMark, achiev, adj, affect, assent, bio, cogproc, drives, function, informal},
    y tick label style={font=\small},
    x tick label style={rotate=90, font=\small},
    xlabel={Linguistic cue},
    ylabel style={text width=5cm,align=center, font=\footnotesize},
    xlabel style={font=\footnotesize},
    ylabel={Relative frequency (\%)}
  ]
  \addplot+ [mark=none, bar shift=-0.2cm] table [x=x, y=y, y error=ey] {
    x y ey
1 11.760912750441143 5.38054826928402
2 1.7157278078004143 2.731042931372431
3 2.017216574028593 2.174442758434573
4 1.3624144034595638 0.5839737403754826
5 4.1510480983285145 1.2650690042569614
6 1.5063775759494216 0.7779530975424134
7 1.3741519399023607 0.7790948717973225
8 2.456467095666378 0.9220633951628869
9 1.4002010993535448 1.646107715728782
10 0.635388468754868 0.36135786563939687
11 1.9557388102485123 0.8797667286227002
12 2.2075910425201153 0.9061989949937503
13 6.029659033274077 1.377419440027138
14 10.59532393253207 2.1620458054334346
  };  \label{pgf:a_liwc}
  
  \addplot+ [mark=none] table [x=x, y=y, y error=ey] {
    x y ey
1 8.084785793913525 1.411114143884575
2 0.3124879921665636 0.2756175200691558
3 0.4404760324331753 0.44408555640033015
4 0.5269435580292723 0.2916376322760712
5 4.902926154704306 1.1778771159651331
6 2.1503106030290904 0.9042284744159503
7 2.4516044114615543 0.9529787883697551
8 4.8070473068015085 1.19793499681301
9 0.46921768707482997 0.6518450170433738
10 1.321991518829754 0.853060881489438
11 2.199696283968133 0.9195330430546957
12 4.888600717054498 1.4498144418442953
13 7.670680549593574 1.2775222018369083
14 12.677110601210181 1.975072322841804
  };  \label{pgf:b_liwc}

  \addplot+ [mark=none, bar shift=0.2cm] table [x=x, y=y, y error=ey] {
    x y ey
1 8.723836033603725 2.569851441664735
2 0.16406221354938738 0.19271859029746113
3 0.8070378023004073 0.8753541085360331
4 0.6209655974400556 0.44598255067243375
5 4.8410148443992975 1.3322930122415038
6 1.9712019217943586 0.8803679362258776
7 2.076269342284822 0.9645521629595994
8 4.2803427100934845 1.2529088731453584
9 0.428302090066796 0.7339766973043649
10 0.6855111345482863 0.4943134337142441
11 1.9653111977584556 0.8893555447621395
12 3.8596531299429353 1.2783262945944207
13 6.200098487168146 1.2439922845182734
14 12.150417553661141 2.071435254005001
  }; \label{pgf:c_liwc}
  
\end{axis}
\end{tikzpicture}


\begin{tikzpicture}
    \matrix[matrix of nodes, draw, 
            nodes={anchor=west, font=\footnotesize},
            column sep=0.5em,
            row sep=0em,
            inner sep=2pt, 
            align=left,
            ampersand replacement=\&] at (current bounding box.south) {

   \ref{pgf:a_liwc} Group A  \& \ref{pgf:b_liwc} Group B \& \ref{pgf:c_liwc} Group C \\
    };
  \end{tikzpicture}

    \caption{Usage of linguistic cues across experimental groups (part one).}
    \label{fig:cues_usage1}
\end{figure}
\begin{figure}[htp]
    \centering
\begin{tikzpicture}
\pgfplotsset{
  error bars/.cd,
    y dir=both, y explicit,
    error bar style={draw=black, line width=0.6pt},   
    error mark options={draw=black, line width=0.6pt},
    error mark=|,  
}

\begin{axis}[
    width=\linewidth,
    height=6cm, ybar,
    ymajorgrids, bar width=0.18cm,
    tick pos=lower,
    tick align=outside,
    xmin=0.5, xmax=14.5,
    ymin=0, ymax=20,
    xtick={1,2,3,4,5,6,7,8,9,10,11,12,13,14},
xticklabels={ingest, interrog, leisure, motion, netspeak, number, percept, posemo, prep, relativ, risk, social, space, time},
    y tick label style={font=\small},
    x tick label style={rotate=90, font=\small},
    xlabel={Linguistic cue},
    ylabel style={text width=5cm,align=center, font=\footnotesize},
    xlabel style={font=\footnotesize},
    ylabel={Relative frequency (\%)},
    legend style={
		at={(0.5,-1)},
		anchor=north,
		legend columns=-1,
		/tikz/every even column/.append style={column sep=0.5cm}
	},
  ]
  \addplot+ [mark=none, bar shift=-0.2cm] table [x=x, y=y, y error=ey] {
    x y ey
1 0.4385147736964096 0.27439960381082434
2 2.1632213031491436 1.0223356599944133
3 1.0645633545920872 0.5171007400016987
4 0.8636341176056321 0.5835904061599531
5 9.02288926101883 2.178674149688126
6 3.227290408527442 1.0405683946127064
7 0.9733355407888102 0.7770062347389217
8 2.243436627648989 0.9299031251497847
9 2.55731393144452 0.8414605869231235
10 5.843938163783477 1.41579519570214
11 0.6240169740711039 0.5299232304298411
12 1.3625857044839775 0.6705887743823493
13 3.479048409252835 1.0724100115048643
14 1.4969976049419313 0.7747424029745227
  };
  \addplot+ [mark=none] table [x expr={\thisrow{x}+0}, y=y, y error=ey] {
    x y ey
1 1.2061946546295286 0.8445204145003997
2 4.829286585543308 1.1527788576204294
3 1.9798938555220067 0.7530883049291481
4 0.4794244485673057 0.34527230724950997
5 10.833927769491636 1.8565486832698674
6 3.010674862874443 0.8923664193404797
7 1.0163651824366111 0.6311073970502856
8 4.513800141268629 1.156409997804916
9 2.0392202691177483 0.7813274810480375
10 5.023683045137247 1.3294000486326585
11 2.21528293135436 0.9602810474082148
12 1.8670294612794613 1.0734390368399267
13 2.936032593037215 0.932792173702774
14 1.6081178299602668 0.6576687937785768
  };

  \addplot+ [mark=none, bar shift=0.2cm] table [x=x, y=y, y error=ey] {
    x y ey
1 0.5987330515202033 0.4624508099654169
2 3.8066681749622924 1.1944033969928172
3 1.565805330634278 0.8205025998141389
4 1.041034708755297 0.5283376439654708
5 10.003131755266208 1.9562025886285195
6 3.417413652910557 1.066946364787214
7 0.9102825591775172 0.6168157027903272
8 3.759230788131717 1.1965224251935407
9 1.4810397090010092 0.6123221367550264
10 4.235506770724484 1.132754134203818
11 1.2824761129823048 0.7770850019123197
12 1.8067037654467994 1.0005775198898035
13 2.0393761765164906 0.7875095408490367
14 1.1548750222086135 0.5859186800617848
  };
  
\end{axis}
\end{tikzpicture}

\begin{tikzpicture}
    \matrix[matrix of nodes, draw, 
            nodes={anchor=west, font=\footnotesize},
            column sep=0.5em,
            row sep=0em,
            inner sep=2pt, 
            align=left,
            ampersand replacement=\&] at (current bounding box.south) {

   \ref{pgf:a_liwc} Group A  \& \ref{pgf:b_liwc} Group B \& \ref{pgf:c_liwc} Group C \\
    };
  \end{tikzpicture}
  
    \caption{Usage of linguistic cues across experimental groups (part two).}
    \label{fig:cues_usage2}
\end{figure}
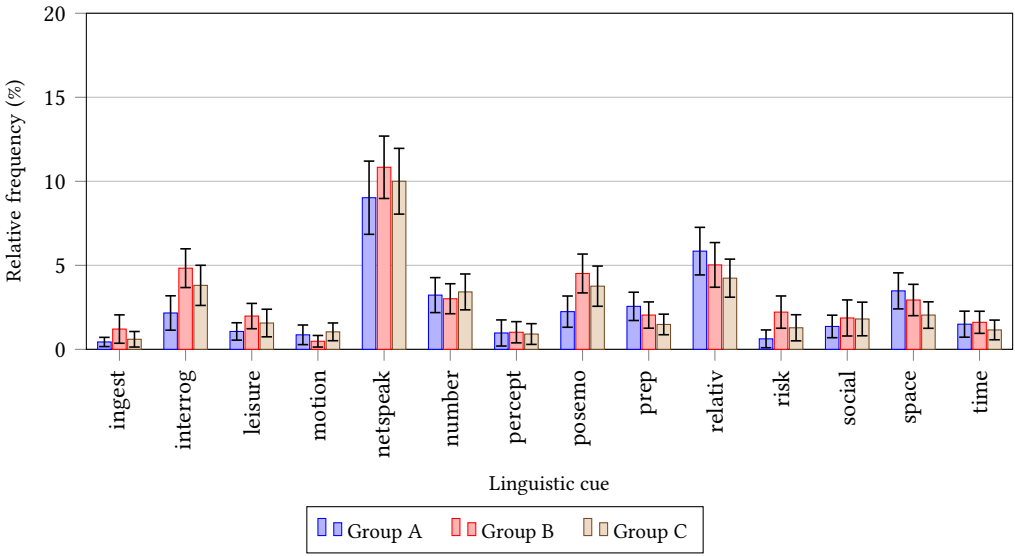

The usage of linguistic cues across the experimental groups is illustrated in \Cref{fig:cues_usage1,fig:cues_usage2}, which depict the relative frequencies of the selected cues together with their 95\% confidence intervals (based on standard errors). As these descriptive intervals substantially overlap, we employed formal non-parametric tests to assess group differences. Specifically, we conducted Kruskal–Wallis tests \citep{Kruskal.1952} for each cue, followed by pairwise Mann–Whitney tests \citep{Mann.1947} with Holm correction for multiple comparisons.




We report the results for all linguistic cues with $p < 0.10$ in \Cref{tbl:kw_posthoc}. The Kruskal–Wallis tests reveal significant differences ($p < 0.05$) across groups in the use of \emph{questions}, \emph{positive emotional expressions}, \emph{motivational language}, and various forms of \emph{punctuation}. Weak differences ($p < 0.10$) are also observed for the use of \emph{spatial references}, \emph{other punctuation marks}, and \emph{prepositions}. Posthoc tests suggest that Group B used more interrogatives, positive emotion words, and motivational concerns, whereas Group A showed higher usage of punctuation marks (periods and dashes), spatial references, and prepositions.


\begin{table}[htp]
    \centering
    \footnotesize
    \caption{Results of Kruskal--Wallis tests and pairwise Mann--Whitney posthoc comparisons (Holm-corrected).}
    \label{tbl:kw_posthoc}
    \begin{tabular}{l
                    S[table-format=2.3,print-zero-integer=false,table-column-width=1.2cm]
                    S[table-format=0.3,print-zero-integer=false,table-column-width=1.2cm]
                    S[table-format=0.3,print-zero-integer=false,table-column-width=1.2cm]
                    p{2cm}}
    \toprule
    Linguistic cue & {$H$} & {$p$} & {$\varepsilon^2$} & Posthoc tests \\
    \midrule
     interrog  & 12.148 & .002 & .036 & A $<$ B ($p=.001$) \\
     Period    & 12.078 & .002 & .036 & A $>$ B ($p=.008$), A $>$ C ($p=.011$) \\
     posemo    &  7.316 & .026 & .019 & A $<$ B ($p=.020$) \\
     Dash      &  6.268 & .044 & .015 & A $>$ C ($p=.050$) \\
     drives    &  6.024 & .049 & .014 & A $<$ B ($p=.043$) \\ \addlinespace
     space     &  5.547 & .062 & .013 & A $>$ C ($p=.054$) \\
     OtherP    &  5.191 & .075 & .011 & n.s. \\
     prep      &  4.952 & .084 & .011 & A $>$ C ($p=.084$) \\
    \bottomrule
    \end{tabular}
    
\end{table}

\subsubsection*{Link of Usage to Subscriptions}

We now analyze whether or not the usage of different linguistic cues across the experimental groups explains subscription and clicking behavior. For this purpose, we estimate logistic regression models of the form
\begin{align}
\log \left(\frac{\Pr(Y = 1)}{\Pr(Y = 0)} \right)
&= \alpha + \beta_{B}\,1_{{\text{Group}=B}}
+ \beta_{C}\,1_{{\text{Group}=C}} + \gamma\,\text{Cue} \\& + \delta_{B}\,1_{{\text{Group}=B}} \times \text{Cue}
+ \delta_{C}\,1_{{\text{Group}=C}} \times \text{Cue},
\label{eq:model_logreg_cues}
\end{align}
where $Y \in \lbrace \text{Subscription}, \text{Click on Affiliated Link} \rbrace$ and Group A serves as the reference.

We then calculate the marginal effects of each linguistic cue on subscriptions and affiliated links for all groups. Standard errors and 95\% confidence intervals are obtained using the variance–covariance matrix of the estimated coefficients (Delta method). 

\begin{figure}[htp]
    \centering
\begin{tikzpicture}
\pgfplotsset{
  error bars/.cd,
    x dir=none,
    y dir=both, y explicit,
}
\begin{axis}[
    width=\linewidth,
    height=6cm, only marks,
    ymajorgrids, 
    tick pos=lower,
    tick align=outside,
    xmin=0.5, xmax=14.5,
    ymin=-0.5, ymax=0.7,
    ytick={0, 0.25, -0.25,0.5, -0.5},
    yticklabels={0, 0.25, $-$0.25,0.50, $-$0.50},
    xtick={1,2,3,4,5,6,7,8,9,10,11,12,13,14},
xticklabels={AllPunc, Dash, OtherP, Period, QMark, achiev, adj, affect, assent, bio, cogproc, drives, function, informal},
    y tick label style={font=\small},
    x tick label style={rotate=90, font=\small},
    xlabel={Linguistic cue},
    ylabel style={text width=5cm,align=center, font=\footnotesize},
    xlabel style={font=\footnotesize},
    ylabel={Marginal effect on subscriptions}
  ]
  
  \addplot+ [mark=o] table [x expr={\thisrow{x}-0.2}, y=y, y error=ey] {
    x y ey
1 0.005089751003380939 0.011207201895635208
2 0.003014945280828487 0.03513346042657485
3 0.029546159518916416 0.030350047925773508
4 0.014404802632121098 0.03643569144376944
5 -0.06831205871494755 0.06650516563294107
6 -0.15183661921689887 0.211034585845059
7 -0.1093102376939077 0.1682931587700782
8 -0.13460585211122453 0.1250252563260231
9 0 0
10 -0.024766283842730306 0.07974078522916225
11 -0.028265273792203192 0.0733668614557377
12 -0.21511884898979947 0.2374383142176215
13 -0.015094601943465139 0.04002393734678944
14 -0.056009278960483126 0.03939052726704544
  };  \label{pgf:marg_a_liwc_sub}
  
  \addplot+ [mark=o] table [x expr={\thisrow{x}+0}, y=y, y error=ey] {
    x y ey
1 0.0028269695236566765 0.021885815119654612
2 -0.017131029407000172 0.09679771683588277
3 0.028291028693896256 0.05476866007006494
4 0.20547719696809144 0.1050264773685313
5 -0.002351948952008742 0.04520621377810564
6 0.10193633585973674 0.057355435284695055
7 0.020186152316136227 0.06687704079798446
8 0.04692765732304208 0.04380174533116757
9 -0.009805951175174243 0.05593415347588008
10 0.10142308839239234 0.04722111442301258
11 0.07236930779826684 0.05488238023225164
12 0.01071328565955193 0.03189832627563251
13 0.02192332772454724 0.031423636755037565
14 -0.006812295391424986 0.020092949955282526
  };  \label{pgf:marg_b_liwc_sub}

  \addplot+ [mark=o] table [x expr={\thisrow{x}+0.2}, y=y, y error=ey] {
    x y ey
1 0.01815503259946347 0.021188251912172485
2 0.4394152459455379 0.2126868247407026
3 0.05509319937822136 0.057610716592887044
4 0.13515246695051292 0.11187098837013976
5 0.006425686958863044 0.055322845532368214
6 -0.013172407415708437 0.09798238084229255
7 0.059554636204313324 0.0735470375165007
8 0.00046718177005330985 0.061807669882347274
9 0.02511220284489113 0.03701552687651167
10 0.14270136892970647 0.0713707714415665
11 0.049640362251407506 0.056344980097518836
12 -0.0327393698240582 0.0739379358000409
13 0.012897991735789743 0.04036701673857947
14 -0.012313516582635464 0.025158805840559004
  }; \label{pgf:marg_c_liwc_sub}
  
\end{axis}
\end{tikzpicture}


\begin{tikzpicture}
    \matrix[matrix of nodes, draw, 
            nodes={anchor=west, font=\footnotesize},
            column sep=0.5em,
            row sep=0em,
            inner sep=2pt, 
            align=left,
            ampersand replacement=\&] at (current bounding box.south) {

   \ref{pgf:marg_a_liwc} Group A  \& \ref{pgf:marg_b_liwc} Group B \& \ref{pgf:marg_c_liwc} Group C \\
    };
  \end{tikzpicture}

    \caption{Marginal effects of linguistic cue usage on subscriptions across experimental groups (part one).}
    \label{fig:cues_marg_sub1}
\end{figure}
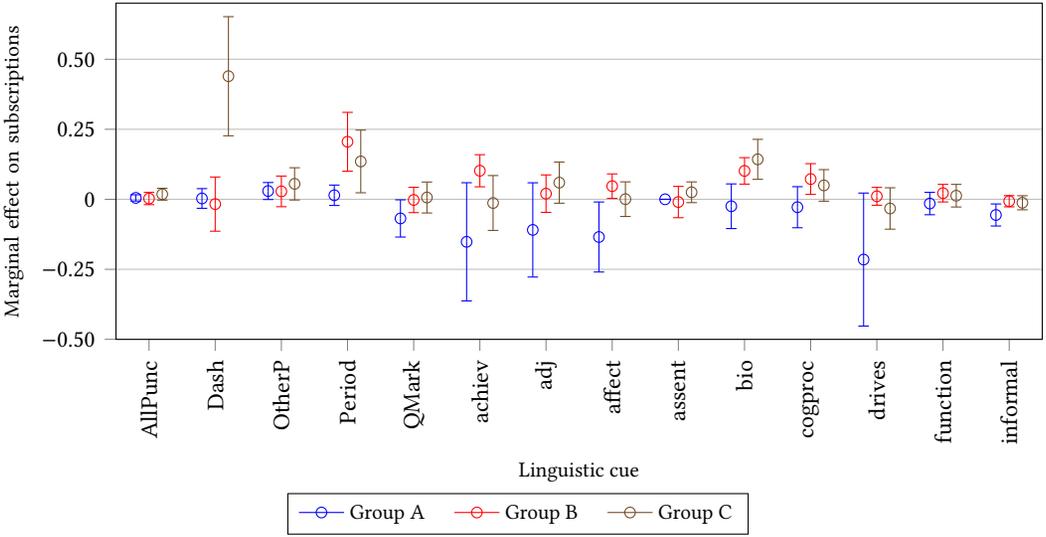
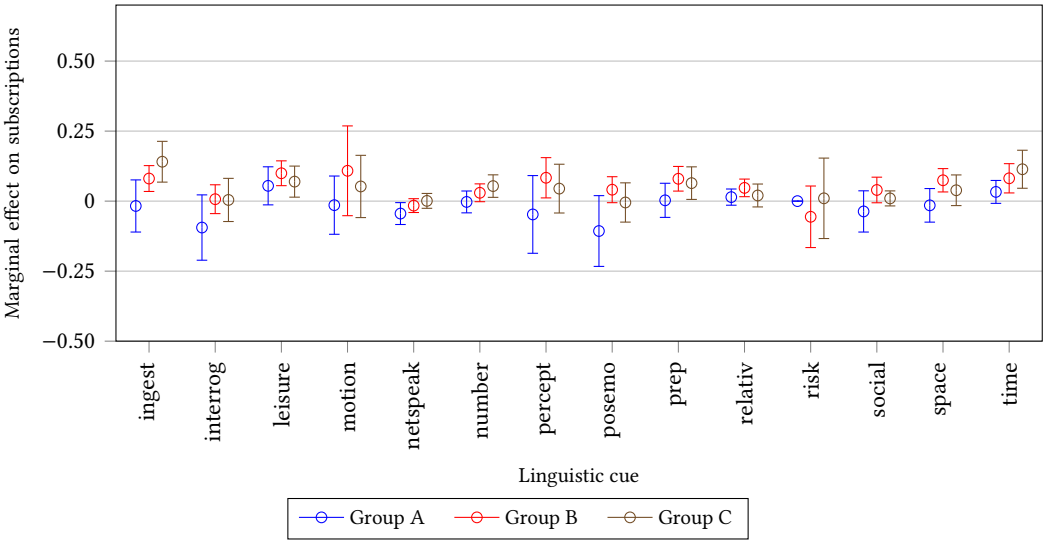
\begin{figure}[htp]
    \centering
\begin{tikzpicture}
\pgfplotsset{
  error bars/.cd,
    x dir=none,
    y dir=both, y explicit,
}
\begin{axis}[
    width=\linewidth,
    height=6cm, only marks,
    ymajorgrids,
    tick pos=lower,
    tick align=outside,
    xmin=0.5, xmax=14.5,
    ymin=-0.5, ymax=0.7,
    ytick={0, 0.25, -0.25,0.5, -0.5},
    yticklabels={0, 0.25, $-$0.25,0.50, $-$0.50},
    xtick={1,2,3,4,5,6,7,8,9,10,11,12,13,14},
xticklabels={ingest, interrog, leisure, motion, netspeak, number, percept, posemo, prep, relativ, risk, social, space, time},
    y tick label style={font=\small},
    x tick label style={rotate=90, font=\small},
    xlabel={Linguistic cue},
    ylabel style={text width=5cm,align=center, font=\footnotesize},
    xlabel style={font=\footnotesize},
    ylabel={Marginal effect on subscriptions}
  ]
  \addplot+ [mark=o] table [x expr={\thisrow{x}-0.2}, y=y, y error=ey] {
    x y ey
1 -0.017321225766770625 0.09308343339766995
2 -0.09435828160153635 0.11665759037685929
3 0.05458565742144348 0.06795629401022898
4 -0.014416821754165091 0.1039871630935515
5 -0.0444203272298694 0.03908228154038665
6 -0.002872630953896324 0.038998334202024774
7 -0.04754593031931193 0.1387337472353332
8 -0.10677397886258669 0.12634841223257692
9 0.00281837832248742 0.060917088137526804
10 0.014305963348944765 0.028935244796024794
11 0 0
12 -0.03681344726704151 0.0736126487469734
13 -0.015093594445176172 0.05994195081897269
14 0.032895539892822574 0.0408917511627668
  };
  \addplot+ [mark=o] table [x expr={\thisrow{x}+0}, y=y, y error=ey] {
    x y ey
1 0.08063145227941773 0.04615014798511144
2 0.006878291700873895 0.051470294707680976
3 0.09948679266318959 0.044309196906343426
4 0.10825751160361496 0.1602905479229566
5 -0.015990128908945356 0.024677014842116703
6 0.029762880696909855 0.03183852824802498
7 0.08337347023998215 0.07187585280766529
8 0.04078148839215576 0.04637273851039808
9 0.07980039242165428 0.04401701883585783
10 0.047256558115807944 0.03142659506059509
11 -0.05594838985749462 0.10986917004264216
12 0.03978054143889327 0.045693612907689155
13 0.07441561119606079 0.04166626146663867
14 0.08148797485387632 0.05224559394930578
  };

  \addplot+ [mark=o] table [x expr={\thisrow{x}+0.2}, y=y, y error=ey] {
    x y ey
1 0.14063422851709415 0.07279926203036671
2 0.004203584837867258 0.077060999809116
3 0.06959755473445083 0.055453442323111626
4 0.052348292757804574 0.11114315584622496
5 0.000862847583951401 0.026537946279719855
6 0.053656339292291105 0.04018979440189713
7 0.044636492444632726 0.08705489618649159
8 -0.00483268663770324 0.07009621159776434
9 0.06422808571258683 0.05815273857330565
10 0.020122803086808316 0.04088571652368832
11 0.01000118422858165 0.14362221350400622
12 0.009681642706009672 0.026820258756883986
13 0.03876949588318526 0.05474397715819579
14 0.1137826172161123 0.06787277875981754
  };
  
\end{axis}
\end{tikzpicture}

\begin{tikzpicture}
    \matrix[matrix of nodes, draw, 
            nodes={anchor=west, font=\footnotesize},
            column sep=0.5em,
            row sep=0em,
            inner sep=2pt, 
            align=left,
            ampersand replacement=\&] at (current bounding box.south) {

  \ref{pgf:marg_a_liwc} Group A  \& \ref{pgf:marg_b_liwc} Group B \& \ref{pgf:marg_c_liwc} Group C \\
    };
  \end{tikzpicture}
  
    \caption{Marginal effects of linguistic cue usage on subscriptions across experimental groups (part two).}
    \label{fig:cues_marg_sub2}
\end{figure}

The results for subscriptions are shown in \Cref{fig:cues_marg_sub1,fig:cues_marg_sub2}. The figures plot the estimated marginal effects of the linguistic cues, along with the 95\% confidence intervals. 
The linguistic cues linked to subscriptions varied across groups. 


In \textbf{Group B}, several categories of content words showed positive relationships. 
For example, users who used more \emph{achievement-related words} were more likely to subscribe  
(\(\text{ME} = 0.10\), 95\% CI [0.04, 0.16]), as were those employing more \emph{biological} terms 
(\(\text{ME} = 0.10\), 95\% CI [0.05, 0.15]) or \emph{cognitive process words} 
(\(\text{ME} = 0.07\), 95\% CI [0.02, 0.13]). 
Effects also extended to structural language, including \emph{prepositions} 
(\(\text{ME} = 0.08\), 95\% CI [0.04, 0.12]) and \emph{time expressions} 
(\(\text{ME} = 0.08\), 95\% CI [0.03, 0.13]). 
These results suggest that users in Group B who wrote in a more analytical and structured style, 
and who referenced goals or biological needs, were more likely to subscribe to the service.  

In \textbf{Group C}, stylistic and topical cues were particularly salient. 
The strongest effect was found for \emph{dash punctuation} 
(\(\text{ME} = 0.44\), 95\% CI [0.23, 0.65]), 
indicating that users who wrote in a more fragmented style were considerably more likely to subscribe. 
Positive effects also appeared for \emph{leisure-related words} 
(\(\text{ME} = 0.07\), 95\% CI [0.01, 0.13]) and \emph{numbers} 
(\(\text{ME} = 0.05\), 95\% CI [0.01, 0.09]), alongside further increases for \emph{bio} and \emph{ingest} terms. 
This pattern suggests that users in Group C who focused on concrete activities, 
quantities, and lifestyle references were more inclined to subscribe.  

By contrast, \textbf{Group A} displayed several negative associations. 
Users who relied more heavily on \emph{question marks} 
(\(\text{ME} = -0.07\), 95\% CI [$-$0.13, $-$0.00]) or \emph{affect words} 
(\(\text{ME} = -0.13\), 95\% CI [$-$0.26, $-$0.01]) were less likely to subscribe. 
Similarly, the use of \emph{informal language} 
(\(\text{ME} = -0.06\), 95\% CI [$-$0.10, $-$0.02]) and \emph{netspeak} 
(\(\text{ME} = -0.04\), 95\% CI [$-$0.08, $-$0.01]) was negatively associated with purchase behavior. 
This indicates that in Group A, a more casual and emotional communication style 
reduced the likelihood of subscriptions to the service.


\subsubsection*{Link of Usage to Clicks on Affiliated Links}

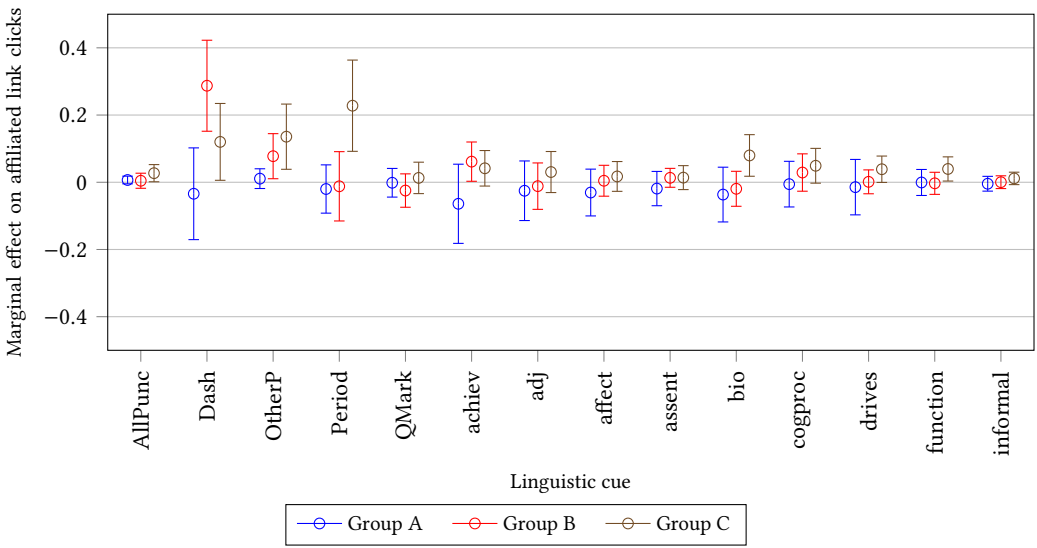
\begin{figure}[htp]
    \centering
\begin{tikzpicture}
\pgfplotsset{
  error bars/.cd,
    x dir=none,
    y dir=both, y explicit,
}
\begin{axis}[
    width=\linewidth,
    height=6cm, only marks,
    ymajorgrids, 
    tick pos=lower,
    tick align=outside,
    xmin=0.5, xmax=14.5,
    ymin=-0.5, ymax=0.5,
    xtick={1,2,3,4,5,6,7,8,9,10,11,12,13,14},
xticklabels={AllPunc, Dash, OtherP, Period, QMark, achiev, adj, affect, assent, bio, cogproc, drives, function, informal},
    y tick label style={font=\small},
    x tick label style={rotate=90, font=\small},
    xlabel={Linguistic cue},
    ylabel style={text width=5cm,align=center, font=\footnotesize},
    xlabel style={font=\footnotesize},
    ylabel={Marginal effect on affiliated link clicks},
  ]
  \addplot+ [mark=o] table [x expr={\thisrow{x}-0.2}, y=y, y error=ey] {
    x y ey
1 0.006762798731295379 0.011394795796396375
2 -0.034189021412014985 0.13649062890126304
3 0.010771337580577607 0.029368225359787857
4 -0.020056847220317982 0.07183729962584905
5 -0.00149068512610113 0.04264058032333125
6 -0.06405578732944293 0.1177956891110468
7 -0.025255689014229858 0.08865566628435856
8 -0.030549303051047368 0.06960858435006002
9 -0.01876925921225259 0.051027604407139784
10 -0.03661378316155709 0.081440478394805
11 -0.005570341836319391 0.06790043990432043
12 -0.01451976216972846 0.08245261504695889
13 -0.0006169760796614893 0.03868419963540968
14 -0.004383058248524609 0.021999638142758342
  };  \label{pgf:marg_a_liwc}
  
  \addplot+ [mark=o] table [x expr={\thisrow{x}}, y=y, y error=ey] {
    x y ey
1 0.004511787068009733 0.022430127326493254
2 0.28721186189428344 0.1354015406182681
3 0.07758962318503929 0.06705755191070062
4 -0.01203453585758445 0.10311232265399768
5 -0.024665272281364687 0.04967590855927072
6 0.06136590648588862 0.058438963803351326
7 -0.01154634479740818 0.06905238182623835
8 0.004521487680666576 0.045957240596888785
9 0.013233190485547826 0.028127797073970884
10 -0.019424412762410144 0.05204080249926509
11 0.028960503615743 0.05559548699083096
12 0.0012894145299307564 0.03552953783839129
13 -0.0031504191928732627 0.03285097104655978
14 0.0002464760152612461 0.019004125027730266
  };  \label{pgf:marg_b_liwc}

  \addplot+ [mark=o] table [x expr={\thisrow{x}+0.2}, y=y, y error=ey] {
    x y ey
1 0.027028234801530368 0.02556708765310644
2 0.12014821571395756 0.11431325194580916
3 0.13561951619493323 0.09702807185710377
4 0.22780612547372284 0.1357561981873632
5 0.012987304225338616 0.046759247128013305
6 0.04156568362234224 0.0528519952208555
7 0.030437830838565112 0.06107618221242133
8 0.017331403182939124 0.044330014431628637
9 0.013831890332172778 0.03563481100801688
10 0.07977808647174073 0.06188768236348749
11 0.04907111770950151 0.051752067475028474
12 0.038754892252059636 0.03923020264498708
13 0.03959639256534712 0.03600290093147244
14 0.011473930795027818 0.018500941115146736
  }; \label{pgf:marg_c_liwc}
  
\end{axis}
\end{tikzpicture}


\begin{tikzpicture}
    \matrix[matrix of nodes, draw, 
            nodes={anchor=west, font=\footnotesize},
            column sep=0.5em,
            row sep=0em,
            inner sep=2pt, 
            align=left,
            ampersand replacement=\&] at (current bounding box.south) {

   \ref{pgf:marg_a_liwc} Group A  \& \ref{pgf:marg_b_liwc} Group B \& \ref{pgf:marg_c_liwc} Group C \\
    };
  \end{tikzpicture}

    \caption{Marginal effects of linguistic cue usage on affiliated link clicks across experimental groups (part one).}
    \label{fig:cues_marg_link1}
\end{figure}

\begin{figure}[htp]
    \centering
\begin{tikzpicture}
\pgfplotsset{
  error bars/.cd,
    x dir=none,
    y dir=both, y explicit,
}
\begin{axis}[
    width=\linewidth,
    height=6cm, only marks,
    ymajorgrids,
    tick pos=lower,
    tick align=outside,
    xmin=0.5, xmax=14.5,
    ymin=-0.5, ymax=0.5,
    xtick={1,2,3,4,5,6,7,8,9,10,11,12,13,14},
xticklabels={ingest, interrog, leisure, motion, netspeak, number, percept, posemo, prep, relativ, risk, social, space, time},
    y tick label style={font=\small},
    x tick label style={rotate=90, font=\small},
    xlabel={Linguistic cue},
    ylabel style={text width=5cm,align=center, font=\footnotesize},
    xlabel style={font=\footnotesize},
    ylabel={Marginal effect on affiliated link clicks},
  ]
  \addplot+ [mark=o] table [x expr={\thisrow{x}-0.2}, y=y, y error=ey] {
    x y ey
1 -0.014174947244881267 0.0835099103656201
2 0.013494264277493019 0.05687679588656311
3 0.009152385256697211 0.07222253046652627
4 0.07390510662253993 0.08003873609217119
5 -0.0028930537516675616 0.021674323767923883
6 -0.06576960317693027 0.0693162555059939
7 0.00925702938008408 0.10268917511194842
8 -0.015053411366192518 0.06495905730428797
9 -0.038263885807998735 0.07106397122553883
10 0.022993989630886624 0.03082111112064476
11 0.06961832677317159 0.12634443802682463
12 -0.06320929464243986 0.08827416760792137
13 -0.03367897232135744 0.06576886416388128
14 0.02416081962145942 0.03549000486054751
  };
  \addplot+ [mark=o] table [x expr={\thisrow{x}+0}, y=y, y error=ey] {
    x y ey
1 -0.005276698681480063 0.04989874683054267
2 0.01763463282902138 0.04681403147563554
3 0.023326719890251913 0.039355533032928644
4 0.1326175149577718 0.15298798219027593
5 -0.0001004858746883777 0.021450148270375672
6 -0.01211546891285882 0.035721791135817665
7 0.01960130708333786 0.07098373658751232
8 -0.004133393166417093 0.04815849426450036
9 0.027465044746295353 0.04361691769774431
10 0.021567141910913794 0.032156643979626164
11 -0.036531532946040085 0.09533815310157452
12 -0.009717388448719201 0.044293011067762955
13 0.02860456516794569 0.040431591776153096
14 -0.008956372166012389 0.055899768738887426
  };

  \addplot+ [mark=o] table [x expr={\thisrow{x}+0.2}, y=y, y error=ey] {
    x y ey
1 0.07886543557780817 0.06306113331627766
2 0.0450688638235702 0.04908242353837672
3 0.06475799154949768 0.05614489968914255
4 -0.013448755134816817 0.1026239738742006
5 0.012129259609720564 0.023376044551008016
6 0.017813348722588135 0.036620313834631865
7 0.0075586116203081545 0.0770765338188888
8 0.02503364650131055 0.04628665128392564
9 0.14072366457994512 0.06551403916962376
10 0.03956598378754514 0.03720067638892062
11 0.026472655637261457 0.09803520054492124
12 0.06000731906316105 0.05419017241513206
13 0.08458814981183568 0.051351036965970515
14 0.08706635934384493 0.06290410929549367
  };
  
\end{axis}
\end{tikzpicture}

\begin{tikzpicture}
    \matrix[matrix of nodes, draw, 
            nodes={anchor=west, font=\footnotesize},
            column sep=0.5em,
            row sep=0em,
            inner sep=2pt, 
            align=left,
            ampersand replacement=\&] at (current bounding box.south) {

  \ref{pgf:marg_a_liwc} Group A  \& \ref{pgf:marg_b_liwc} Group B \& \ref{pgf:marg_c_liwc} Group C \\
    };
  \end{tikzpicture}
  
    \caption{Marginal effects of linguistic cue usage on affiliated link clicks across experimental groups (part two).}
    \label{fig:cues_marg_link2}
\end{figure}
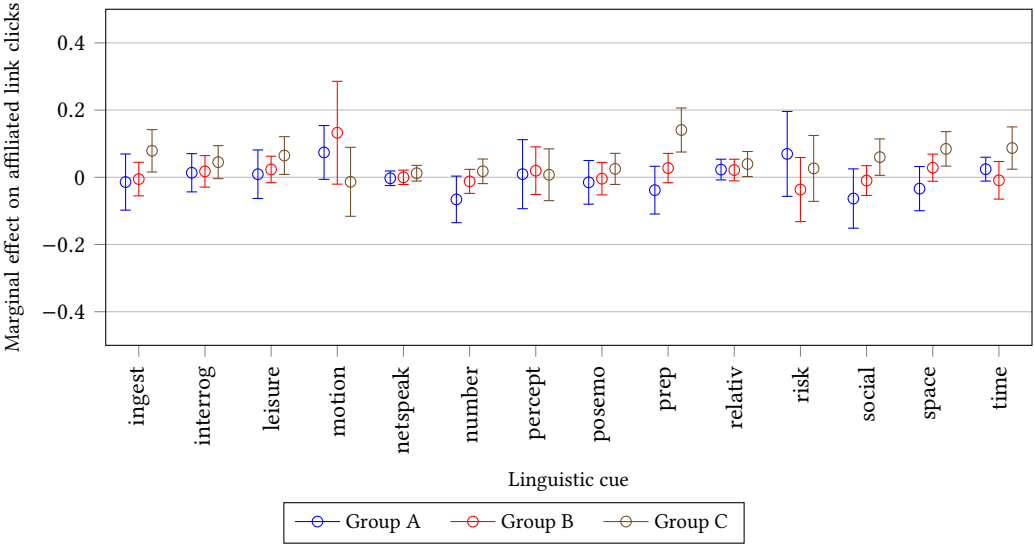

We also aim to understand the link between the usage of linguistic cues and clicks on affiliated links. The estimated marginal effects along with the 95\% confidence intervals are shown in \Cref{fig:cues_marg_link1,fig:cues_marg_link2}. Here, significant patterns again emerged varying by group. 

In \textbf{Group B}, users who relied more on \emph{dash punctuation} were substantially more likely to click 
(\(\text{ME} = 0.29\), 95\% CI [0.15, 0.42]), and additional positive effects were observed for 
\emph{achievement words} (\(\text{ME} = 0.06\), 95\% CI [0.00, 0.12]) and \emph{other punctuation} 
(\(\text{ME} = 0.08\), 95\% CI [0.01, 0.14]). 
This suggests that fragmented sentence structure and goal-oriented language were associated with stronger click behavior.  

In \textbf{Group C}, a broader set of cues was predictive. 
Positive effects were observed for multiple categories of punctuation, including \emph{all punctuation} 
(\(\text{ME} = 0.03\), 95\% CI [0.00, 0.05]), \emph{dash punctuation} 
(\(\text{ME} = 0.12\), 95\% CI [0.01, 0.23]), \emph{other punctuation} 
(\(\text{ME} = 0.14\), 95\% CI [0.04, 0.23]), and \emph{periods} 
(\(\text{ME} = 0.23\), 95\% CI [0.09, 0.36]). 
In addition, \emph{bio} (\(\text{ME} = 0.08\), 95\% CI [0.02, 0.14]) and \emph{ingest words} 
(\(\text{ME} = 0.08\), 95\% CI [0.02, 0.14]) as well as references to \emph{leisure} 
(\(\text{ME} = 0.06\), 95\% CI [0.01, 0.12]) predicted higher click likelihood. 
Further linguistic markers such as \emph{prepositions} (\(\text{ME} = 0.14\), 95\% CI [0.08, 0.21]), 
\emph{relativity} (\(\text{ME} = 0.04\), 95\% CI [0.00, 0.08]), 
\emph{social words} (\(\text{ME} = 0.06\), 95\% CI [0.01, 0.11]), 
\emph{space} (\(\text{ME} = 0.08\), 95\% CI [0.03, 0.14]), 
and \emph{time} (\(\text{ME} = 0.09\), 95\% CI [0.02, 0.15]) were also positively related to clicking. 
This indicates that Group C users who expressed themselves in a more descriptive, socially anchored, 
and temporally specific manner showed higher engagement through link clicks.  

By contrast, \textbf{Group A} exhibited no significant associations between linguistic cues and clicking behavior.


\section{Discussion}
\label{sec:discussion}

\subsection{Summary of Findings}

Our field experiment demonstrates that the communication tone of GenAI significantly influences user engagement and purchase behavior in online travel planning. Specifically, users interacting with an enthusiastic chatbot (Group A) produced longer prompts, while both enthusiastic (Group A) and neutral tones (Group B) increased subscription likelihood compared to the control group (C) with no tone instructions. Although prompt quantity did not vary across conditions, linguistic analyses revealed systematic differences in how users expressed themselves, with positive emotion, interrogatives, and motivational language more prevalent in the neutral condition, and punctuation, spatial references, and prepositions more frequent in the enthusiastic condition. Importantly, several linguistic cues were found to predict subscription and click behaviors, suggesting that user communication style mediates the persuasive effects of AI tone. 

Here, we found interesting interactions between the tone of GenAI and linguistic cues in user prompts. In Group A, a more casual and emotional communication style reduced the likelihood of subscriptions to the service. Similarly, users in Group B who wrote in a more analytical and structured style, and who referenced goals or biological needs, were more likely to subscribe. By contrast, users in Group C who wrote in a more fragmented way, using
quantifiers, and lifestyle references were more inclined to subscribe. Regarding clicks on affiliated links, linguistic cues had no significant associations for users in Group A. Users in Group B communicating in a more fragmented sentence structure and more goal-oriented language were associated with stronger click behavior. Higher click rates were observed for Group C users who expressed themselves in a more descriptive, socially anchored, and temporally specific manner.


\subsection{Theoretical Implications}


To the best of our knowledge, our study provides the first field experimental evidence on the use of GenAI for travel itinerary planning in a consumer-facing context. Prior work has primarily focused on prototypes \citep{Ju.2024,Udandarao.2025} or benchmark studies \citep{Xie.2024,Shen.2025} that highlight technical feasibility and inherent challenges such as hallucinations. By contrast, we demonstrate in a real-world deployment with paying users that GenAI can indeed be applied in itinerary planning. In particular, we find that success does not depend solely on technical performance but is strongly shaped by design choices such as the tone instructions of the GenAI. These findings suggest that theories of GenAI use in complex planning tasks should not only focus on technical dimensions (e.g., accuracy, feasibility, constraint handling) but also integrate social and communicative aspects of design to fully explain user acceptance and impact.


In addition, our study extends prior research that has used linguistic cues to capture aspects of user experience \citep[e.g.,][]{Jang.2017,Fang.2025}. We measure such cues in user prompts through LIWC and show that they are linked to concrete purchase behavior. While these results are exploratory in nature and should be interpreted with caution, they nonetheless indicate systematic patterns that link linguistic expression to consumer decisions. Taken together, these findings suggest that linguistic signals can serve as a bridge between the experiential dimension of user interaction and actual consumer decisions, highlighting the importance of integrating linguistic cues into theories of GenAI persuasion and adoption. For example, future systems could dynamically adapt their tone when users employ highly motivational or emotionally positive language, or provide additional informational support when user prompts contain frequent interrogatives. More broadly, linguistic analysis could help detect moments when users are still exploring versus ready to commit, enabling GenAI to calibrate its communication style in real time.




Our findings also contribute to the broader discussion of AI in customer service by highlighting the trade-off between warmth and neutrality in GenAI communication. In our study, an enthusiastic (warm) tone proved effective in stimulating engagement and shaping linguistic expression, consistent with prior work emphasizing the relational benefits of warmth in service interactions \citep{Kull.2021}. At the same time, a neutral tone was equally effective in increasing subscription likelihood, suggesting that neutrality can enhance perceptions of competence and credibility, especially in cognitively demanding planning tasks. These results support the notion that the expected language of AI agents is context-dependent \citep{Chaves.2020}: while warmth may be advantageous in affect-rich and exploratory settings, neutrality may be better suited for tasks that require focus and rational evaluation. Future research should further examine this contextual dependency and explore to what extent our findings generalize beyond pleasurable tasks such as holiday planning to more critical or stressful customer service situations.


\subsection{Practical Implications}

The findings offer several implications for the design of consumer-facing GenAI systems in online travel planning. Incorporating explicit tone instructions is preferable to leaving the model unguided, as both enthusiastic and neutral tones increase subscription likelihood compared to the control. While enthusiastic framing stimulated longer and more detailed user prompts, neutral framing encouraged more structured and goal-oriented communication, both of which were linked to higher subscription probability. 

Notably, industry guidelines such as Microsoft Azure’s documentation \citep{Microsoft} recommend specifying tone of voice as a best practice when prompting GPT models, yet provide little clarity on how to define or select an appropriate tone in practice; identifying effective styles thus requires deliberate design choices and empirical testing. 
From a design perspective, chatbot agent tone should be calibrated to foster engagement without undermining perceived competence. Tone can be leveraged as a design lever not only to enhance user experience but also to guide economic outcomes in GenAI-enabled travel platforms.


\subsection{Limitations and Future Research}

Our study has several limitations that warrant consideration. First, the field experiment was conducted within a single commercial platform with users largely from North America, which may limit the generalizability of findings to other domains of GenAI use. Second, the manipulation of tone was restricted to enthusiastic versus neutral framing, leaving open whether alternative tones (e.g., formal, empathetic) would yield different effects. Third, outcomes such as subscription and affiliate clicks capture short-term behaviors, but longer-term impacts of tone on sustained engagement (e.g., actual booking of the planned itinerary) remain unexplored. 



In the near future, we plan to replicate and extend the findings from the field experiment in a controlled lab experiment. While the field study achieves greater ecological validity \citep{Lew.2011}, a lab study will allow us to measure cognitive and emotional activity using EEG to better understand the driving factors that guide user behavior \citep{Dimoka.2012}. 


\bibliographystyle{ACM-Reference-Format}
\bibliography{literature}


\begin{thebibliography}{54}


\ifx \showCODEN    \undefined \def \showCODEN     #1{\unskip}     \fi
\ifx \showISBNx    \undefined \def \showISBNx     #1{\unskip}     \fi
\ifx \showISBNxiii \undefined \def \showISBNxiii  #1{\unskip}     \fi
\ifx \showISSN     \undefined \def \showISSN      #1{\unskip}     \fi
\ifx \showLCCN     \undefined \def \showLCCN      #1{\unskip}     \fi
\ifx \shownote     \undefined \def \shownote      #1{#1}          \fi
\ifx \showarticletitle \undefined \def \showarticletitle #1{#1}   \fi
\ifx \showURL      \undefined \def \showURL       {\relax}        \fi
\providecommand\bibfield[2]{#2}
\providecommand\bibinfo[2]{#2}
\providecommand\natexlab[1]{#1}
\providecommand\showeprint[2][]{arXiv:#2}

\bibitem[Adam et~al\mbox{.}(2021)]%
        {Adam.2021}
\bibfield{author}{\bibinfo{person}{Martin Adam}, \bibinfo{person}{Michael
  Wessel}, {and} \bibinfo{person}{Alexander Benlian}.}
  \bibinfo{year}{2021}\natexlab{}.
\newblock \showarticletitle{{AI-based} chatbots in customer service and their
  effects on user compliance}.
\newblock \bibinfo{journal}{\emph{Electronic Markets}} \bibinfo{volume}{31},
  \bibinfo{number}{2} (\bibinfo{year}{2021}), \bibinfo{pages}{427--445}.
\newblock


\bibitem[Azure(2024)]%
        {Microsoft}
\bibfield{author}{\bibinfo{person}{Microsoft Azure}.}
  \bibinfo{year}{2024}\natexlab{}.
\newblock \bibinfo{title}{System Message -- Tone of voice}.
\newblock
\newblock
\shownote{\url{https://microsoft.github.io/Workshop-Interact-with-OpenAI-models/Part-2-labs/System-Message/\#tone-of-voice}}.


\bibitem[Baayen et~al\mbox{.}(2008)]%
        {Baayen.2008}
\bibfield{author}{\bibinfo{person}{R~Harald Baayen}, \bibinfo{person}{Douglas~J
  Davidson}, {and} \bibinfo{person}{Douglas~M Bates}.}
  \bibinfo{year}{2008}\natexlab{}.
\newblock \showarticletitle{Mixed-effects modeling with crossed random effects
  for subjects and items}.
\newblock \bibinfo{journal}{\emph{Journal of Memory and Language}}
  \bibinfo{volume}{59}, \bibinfo{number}{4} (\bibinfo{year}{2008}),
  \bibinfo{pages}{390--412}.
\newblock


\bibitem[Banerjee et~al\mbox{.}(2025)]%
        {Banerjee.2025}
\bibfield{author}{\bibinfo{person}{Ashmi Banerjee}, \bibinfo{person}{Adithi
  Satish}, \bibinfo{person}{Fitri~Nur Aisyah}, \bibinfo{person}{Wolfgang
  W{\"o}rndl}, {and} \bibinfo{person}{Yashar Deldjoo}.}
  \bibinfo{year}{2025}\natexlab{}.
\newblock \showarticletitle{SynthTRIPs: A Knowledge-Grounded Framework for
  Benchmark Data Generation for Personalized Tourism Recommenders}. In
  \bibinfo{booktitle}{\emph{Proceedings of the 48th International ACM SIGIR
  Conference on Research and Development in Information Retrieval}}.
  \bibinfo{pages}{3743--3752}.
\newblock


\bibitem[Bolzoni et~al\mbox{.}(2014)]%
        {Bolzoni.2014}
\bibfield{author}{\bibinfo{person}{Paolo Bolzoni}, \bibinfo{person}{Sven
  Helmer}, \bibinfo{person}{Kevin Wellenzohn}, \bibinfo{person}{Johann Gamper},
  {and} \bibinfo{person}{Periklis Andritsos}.} \bibinfo{year}{2014}\natexlab{}.
\newblock \showarticletitle{Efficient itinerary planning with category
  constraints}. In \bibinfo{booktitle}{\emph{Proceedings of the 22nd ACM
  SIGSPATIAL International Conference on Advances in Geographic Information
  Systems}}. \bibinfo{pages}{203--212}.
\newblock


\bibitem[Burgoon(1995)]%
        {Burgoon.1995.LET}
\bibfield{author}{\bibinfo{person}{Michael Burgoon}.}
  \bibinfo{year}{1995}\natexlab{}.
\newblock \showarticletitle{Language expectancy theory: Elaboration,
  explication, and extension}.
\newblock In \bibinfo{booktitle}{\emph{{Communication and Social Influence
  Processes}}}, \bibfield{editor}{\bibinfo{person}{C~R Berger} {and}
  \bibinfo{person}{M.~Burgoon}} (Eds.). \bibinfo{publisher}{Michigan State
  University Press}, \bibinfo{address}{East Lansing}, \bibinfo{pages}{29--52}.
\newblock


\bibitem[Burgoon and Miller(1985)]%
        {Burgoon.1985.LET}
\bibfield{author}{\bibinfo{person}{Michael Burgoon} {and}
  \bibinfo{person}{Gerald~R Miller}.} \bibinfo{year}{1985}\natexlab{}.
\newblock \showarticletitle{An expectancy interpretation of language and
  persuasion}.
\newblock In \bibinfo{booktitle}{\emph{{Recent Advances in Language,
  Communication, and Social Psychology}}},
  \bibfield{editor}{\bibinfo{person}{H~Giles} {and} \bibinfo{person}{R~N
  St.~Clair}} (Eds.). \bibinfo{publisher}{Wiley \& Sons},
  \bibinfo{address}{London}, \bibinfo{pages}{51--72}.
\newblock


\bibitem[Chaves(2020)]%
        {Chaves.2020}
\bibfield{author}{\bibinfo{person}{Ana~Paula Chaves}.}
  \bibinfo{year}{2020}\natexlab{}.
\newblock \showarticletitle{Should my chatbot be register-specific? {D}esigning
  appropriate utterances for tourism}. In \bibinfo{booktitle}{\emph{Extended
  Abstracts of the CHI Conference on Human Factors in Computing Systems}}.
  \bibinfo{pages}{1--11}.
\newblock


\bibitem[Diederich et~al\mbox{.}(2022)]%
        {Diederich.2022}
\bibfield{author}{\bibinfo{person}{Stephan Diederich},
  \bibinfo{person}{Alfred~Benedikt Brendel}, \bibinfo{person}{Stefan Morana},
  {and} \bibinfo{person}{Lutz Kolbe}.} \bibinfo{year}{2022}\natexlab{}.
\newblock \showarticletitle{On the design of and interaction with
  conversational agents: {A}n organizing and assessing review of human-computer
  interaction research}.
\newblock \bibinfo{journal}{\emph{Journal of the Association for Information
  Systems}} \bibinfo{volume}{23}, \bibinfo{number}{1} (\bibinfo{year}{2022}),
  \bibinfo{pages}{96--138}.
\newblock


\bibitem[Dimoka et~al\mbox{.}(2012)]%
        {Dimoka.2012}
\bibfield{author}{\bibinfo{person}{Angelika Dimoka}, \bibinfo{person}{Rajiv~D.
  Banker}, \bibinfo{person}{Izak Benbasat}, \bibinfo{person}{Fred~D. Davis},
  \bibinfo{person}{Alan~R. Dennis}, \bibinfo{person}{David Gefen},
  \bibinfo{person}{Alok Gupta}, \bibinfo{person}{Anja Ischebeck},
  \bibinfo{person}{Peter~H. Kenning}, \bibinfo{person}{Paul~A. Pavlou},
  {et~al\mbox{.}}} \bibinfo{year}{2012}\natexlab{}.
\newblock \showarticletitle{On the Use of Neurophysiological Tools in
  {IS}~Research: Developing a Research Agenda for {NeuroIS}}.
\newblock \bibinfo{journal}{\emph{MIS Quarterly}} \bibinfo{volume}{36},
  \bibinfo{number}{3} (\bibinfo{year}{2012}), \bibinfo{pages}{679--702}.
\newblock


\bibitem[D'Mello and Calvo(2013)]%
        {Mello.2013}
\bibfield{author}{\bibinfo{person}{Sidney D'Mello} {and}
  \bibinfo{person}{Rafael~A Calvo}.} \bibinfo{year}{2013}\natexlab{}.
\newblock \showarticletitle{Beyond the basic emotions: {W}hat should affective
  computing compute?}
\newblock In \bibinfo{booktitle}{\emph{Extended Abstracts of the CHI Conference
  on Human Factors in Computing Systems}}. \bibinfo{pages}{2287--2294}.
\newblock


\bibitem[Fang and Zhu(2023)]%
        {Fang.2023}
\bibfield{author}{\bibinfo{person}{Anna Fang} {and} \bibinfo{person}{Haiyi
  Zhu}.} \bibinfo{year}{2023}\natexlab{}.
\newblock \showarticletitle{Measuring the stigmatizing effects of a highly
  publicized event on online mental health discourse}. In
  \bibinfo{booktitle}{\emph{Proceedings of the 2023 CHI Conference on Human
  Factors in Computing Systems}}. \bibinfo{pages}{1--18}.
\newblock


\bibitem[Fang et~al\mbox{.}(2025)]%
        {Fang.2025}
\bibfield{author}{\bibinfo{person}{Cathy~Mengying Fang},
  \bibinfo{person}{Auren~R. Liu}, \bibinfo{person}{Valdemar Danry},
  \bibinfo{person}{Eunhae Lee}, \bibinfo{person}{Samantha W.~T. Chan},
  \bibinfo{person}{Pat Pataranutaporn}, \bibinfo{person}{Pattie Maes},
  \bibinfo{person}{Jason Phang}, \bibinfo{person}{Michael Lampe},
  \bibinfo{person}{Lama Ahmad}, {and} \bibinfo{person}{Sandhini Agarwal}.}
  \bibinfo{year}{2025}\natexlab{}.
\newblock \bibinfo{title}{How AI and Human Behaviors Shape Psychosocial Effects
  of Chatbot Use: A Longitudinal Randomized Controlled Study}.
\newblock
\showeprint[arxiv]{2503.17473}~[cs.HC]
\urldef\tempurl%
\url{https://arxiv.org/abs/2503.17473}
\showURL{%
\tempurl}


\bibitem[Ferland et~al\mbox{.}(2025)]%
        {Ferland.2025}
\bibfield{author}{\bibinfo{person}{Libby Ferland}, \bibinfo{person}{Risako
  Owan}, \bibinfo{person}{Zachary Kunkel}, \bibinfo{person}{Hannah Qu},
  \bibinfo{person}{Maria Gini}, {and} \bibinfo{person}{Wilma Koutstaal}.}
  \bibinfo{year}{2025}\natexlab{}.
\newblock \showarticletitle{What are you looking forward to? {D}eliberate
  positivity as a promising strategy for conversational agents}.
\newblock \bibinfo{journal}{\emph{ACM Transactions on Interactive Intelligent
  Systems}} (\bibinfo{year}{2025}).
\newblock


\bibitem[Feuerriegel et~al\mbox{.}(2024)]%
        {Feuerriegel.2024}
\bibfield{author}{\bibinfo{person}{Stefan Feuerriegel}, \bibinfo{person}{Jochen
  Hartmann}, \bibinfo{person}{Christian Janiesch}, {and}
  \bibinfo{person}{Patrick Zschech}.} \bibinfo{year}{2024}\natexlab{}.
\newblock \showarticletitle{Generative {AI}}.
\newblock \bibinfo{journal}{\emph{Business \& Information Systems Engineering}}
  \bibinfo{volume}{66}, \bibinfo{number}{1} (\bibinfo{year}{2024}),
  \bibinfo{pages}{111--126}.
\newblock


\bibitem[Gnewuch et~al\mbox{.}(2024)]%
        {Gnewuch.2024}
\bibfield{author}{\bibinfo{person}{Ulrich Gnewuch}, \bibinfo{person}{Stefan
  Morana}, \bibinfo{person}{Oliver Hinz}, \bibinfo{person}{Ralf Kellner}, {and}
  \bibinfo{person}{Alexander Maedche}.} \bibinfo{year}{2024}\natexlab{}.
\newblock \showarticletitle{More than a bot? {T}he impact of disclosing human
  involvement on customer interactions with hybrid service agents}.
\newblock \bibinfo{journal}{\emph{Information Systems Research}}
  \bibinfo{volume}{35}, \bibinfo{number}{3} (\bibinfo{year}{2024}),
  \bibinfo{pages}{936--955}.
\newblock


\bibitem[Gnewuch et~al\mbox{.}(2017)]%
        {Gnewuch.2017}
\bibfield{author}{\bibinfo{person}{Ulrich Gnewuch}, \bibinfo{person}{Stefan
  Morana}, {and} \bibinfo{person}{Alexander Maedche}.}
  \bibinfo{year}{2017}\natexlab{}.
\newblock \showarticletitle{Towards Designing Cooperative and Social
  Conversational Agents for Customer Service.}. In
  \bibinfo{booktitle}{\emph{International Conference on Information Systems}}.
  \bibinfo{pages}{1--13}.
\newblock


\bibitem[Gray and Wegner(2012)]%
        {Gray.2012}
\bibfield{author}{\bibinfo{person}{Kurt Gray} {and} \bibinfo{person}{Daniel~M
  Wegner}.} \bibinfo{year}{2012}\natexlab{}.
\newblock \showarticletitle{Feeling robots and human zombies: Mind perception
  and the uncanny valley}.
\newblock \bibinfo{journal}{\emph{Cognition}} \bibinfo{volume}{125},
  \bibinfo{number}{1} (\bibinfo{year}{2012}), \bibinfo{pages}{125--130}.
\newblock


\bibitem[Gremler and Gwinner(2000)]%
        {Gremler.2020}
\bibfield{author}{\bibinfo{person}{Dwayne~D Gremler} {and}
  \bibinfo{person}{Kevin~P Gwinner}.} \bibinfo{year}{2000}\natexlab{}.
\newblock \showarticletitle{Customer-employee rapport in service
  relationships}.
\newblock \bibinfo{journal}{\emph{Journal of Service Research}}
  \bibinfo{volume}{3}, \bibinfo{number}{1} (\bibinfo{year}{2000}),
  \bibinfo{pages}{82--104}.
\newblock


\bibitem[Han et~al\mbox{.}(2023)]%
        {han.2023}
\bibfield{author}{\bibinfo{person}{Elizabeth Han}, \bibinfo{person}{Dezhi Yin},
  {and} \bibinfo{person}{Han Zhang}.} \bibinfo{year}{2023}\natexlab{}.
\newblock \showarticletitle{Bots with feelings: {S}hould {AI} agents express
  positive emotion in customer service?}
\newblock \bibinfo{journal}{\emph{Information Systems Research}}
  \bibinfo{volume}{34}, \bibinfo{number}{3} (\bibinfo{year}{2023}),
  \bibinfo{pages}{1296--1311}.
\newblock


\bibitem[He et~al\mbox{.}(2025)]%
        {He.2025}
\bibfield{author}{\bibinfo{person}{Gaole He}, \bibinfo{person}{Gianluca
  Demartini}, {and} \bibinfo{person}{Ujwal Gadiraju}.}
  \bibinfo{year}{2025}\natexlab{}.
\newblock \showarticletitle{Plan-then-execute: {A}n empirical study of user
  trust and team performance when using {LLM} agents as a daily assistant}. In
  \bibinfo{booktitle}{\emph{Proceedings of the 2025 CHI Conference on Human
  Factors in Computing Systems}}. \bibinfo{pages}{1--22}.
\newblock


\bibitem[He et~al\mbox{.}(2024)]%
        {He.2024}
\bibfield{author}{\bibinfo{person}{Linwei He}, \bibinfo{person}{Anouck
  Braggaar}, \bibinfo{person}{Erkan Basar}, \bibinfo{person}{Emiel Krahmer},
  \bibinfo{person}{Marjolijn Antheunis}, {and} \bibinfo{person}{Reinout
  Wiers}.} \bibinfo{year}{2024}\natexlab{}.
\newblock \showarticletitle{Exploring user engagement through an interaction
  lens: {W}hat textual cues can tell us about human-chatbot interactions}. In
  \bibinfo{booktitle}{\emph{Proceedings of the 6th ACM Conference on
  Conversational User Interfaces}}. \bibinfo{pages}{1--14}.
\newblock


\bibitem[Hu et~al\mbox{.}(2018)]%
        {Hu.2018.CHI}
\bibfield{author}{\bibinfo{person}{Tianran Hu}, \bibinfo{person}{Anbang Xu},
  \bibinfo{person}{Zhe Liu}, \bibinfo{person}{Quanzeng You},
  \bibinfo{person}{Yufan Guo}, \bibinfo{person}{Vibha Sinha},
  \bibinfo{person}{Jiebo Luo}, {and} \bibinfo{person}{Rama Akkiraju}.}
  \bibinfo{year}{2018}\natexlab{}.
\newblock \showarticletitle{Touch your heart: A tone-aware chatbot for customer
  care on social media}. In \bibinfo{booktitle}{\emph{Proceedings of the 2018
  CHI Conference on Human Factors in Computing Systems}}.
  \bibinfo{pages}{1--12}.
\newblock


\bibitem[Jang and Yi(2017)]%
        {Jang.2017}
\bibfield{author}{\bibinfo{person}{Jincheul Jang} {and}
  \bibinfo{person}{Mun~Yong Yi}.} \bibinfo{year}{2017}\natexlab{}.
\newblock \showarticletitle{Modeling user satisfaction from the extraction of
  user experience elements in online product reviews}. In
  \bibinfo{booktitle}{\emph{Extended Abstracts of the CHI Conference on Human
  Factors in Computing Systems}}. \bibinfo{pages}{1718--1725}.
\newblock


\bibitem[JU et~al\mbox{.}(2024)]%
        {Ju.2024}
\bibfield{author}{\bibinfo{person}{Da JU}, \bibinfo{person}{Song Jiang},
  \bibinfo{person}{Andrew Cohen}, \bibinfo{person}{Aaron Foss},
  \bibinfo{person}{Sasha Mitts}, \bibinfo{person}{Arman Zharmagambetov},
  \bibinfo{person}{Brandon Amos}, \bibinfo{person}{Xian Li},
  \bibinfo{person}{Justine~T Kao}, \bibinfo{person}{Maryam Fazel-Zarandi},
  {and} \bibinfo{person}{Yuandong Tian}.} \bibinfo{year}{2024}\natexlab{}.
\newblock \bibinfo{title}{To the Globe (TTG): Towards Language-Driven
  Guaranteed Travel Planning}.
\newblock
\showeprint[arxiv]{2410.16456}~[cs.CL]
\urldef\tempurl%
\url{https://arxiv.org/abs/2410.16456}
\showURL{%
\tempurl}


\bibitem[Kranzb{\"u}hler et~al\mbox{.}(2020)]%
        {Kranzbuhler.2020}
\bibfield{author}{\bibinfo{person}{Anne-Madeleine Kranzb{\"u}hler},
  \bibinfo{person}{Alfred Zerres}, \bibinfo{person}{Mirella~HP Kleijnen}, {and}
  \bibinfo{person}{Peeter~WJ Verlegh}.} \bibinfo{year}{2020}\natexlab{}.
\newblock \showarticletitle{Beyond valence: A meta-analysis of discrete
  emotions in firm-customer encounters}.
\newblock \bibinfo{journal}{\emph{Journal of the Academy of Marketing Science}}
   \bibinfo{volume}{48} (\bibinfo{year}{2020}), \bibinfo{pages}{478--498}.
\newblock


\bibitem[Kruskal and Wallis(1952)]%
        {Kruskal.1952}
\bibfield{author}{\bibinfo{person}{William~H Kruskal} {and}
  \bibinfo{person}{W~Allen Wallis}.} \bibinfo{year}{1952}\natexlab{}.
\newblock \showarticletitle{Use of ranks in one-criterion variance analysis}.
\newblock \bibinfo{journal}{\emph{Journal of the American statistical
  Association}} \bibinfo{volume}{47}, \bibinfo{number}{260}
  (\bibinfo{year}{1952}), \bibinfo{pages}{583--621}.
\newblock


\bibitem[Kull et~al\mbox{.}(2021)]%
        {Kull.2021}
\bibfield{author}{\bibinfo{person}{Alexander~J. Kull},
  \bibinfo{person}{Marisabel Romero}, {and} \bibinfo{person}{Lisa Monahan}.}
  \bibinfo{year}{2021}\natexlab{}.
\newblock \showarticletitle{How may {I} help you? {D}riving brand engagement
  through the warmth of an initial chatbot message}.
\newblock \bibinfo{journal}{\emph{Journal of Business Research}}
  \bibinfo{volume}{135} (\bibinfo{year}{2021}), \bibinfo{pages}{840--850}.
\newblock


\bibitem[Lee et~al\mbox{.}(2025)]%
        {Lee.2025}
\bibfield{author}{\bibinfo{person}{Christine~P Lee}, \bibinfo{person}{David
  Porfirio}, \bibinfo{person}{Xinyu~Jessica Wang},
  \bibinfo{person}{Kevin~Chenkai Zhao}, {and} \bibinfo{person}{Bilge Mutlu}.}
  \bibinfo{year}{2025}\natexlab{}.
\newblock \showarticletitle{Veriplan: Integrating formal verification and llms
  into end-user planning}. In \bibinfo{booktitle}{\emph{Proceedings of the 2025
  CHI Conference on Human Factors in Computing Systems}}.
  \bibinfo{pages}{1--19}.
\newblock


\bibitem[Lee and Choi(2017)]%
        {Lee.2017}
\bibfield{author}{\bibinfo{person}{SeoYoung Lee} {and} \bibinfo{person}{Junho
  Choi}.} \bibinfo{year}{2017}\natexlab{}.
\newblock \showarticletitle{Enhancing user experience with conversational agent
  for movie recommendation: {E}ffects of self-disclosure and reciprocity}.
\newblock \bibinfo{journal}{\emph{International Journal of Human-Computer
  Studies}}  \bibinfo{volume}{103} (\bibinfo{year}{2017}),
  \bibinfo{pages}{95--105}.
\newblock


\bibitem[Lew et~al\mbox{.}(2011)]%
        {Lew.2011}
\bibfield{author}{\bibinfo{person}{Letitia Lew}, \bibinfo{person}{Truc Nguyen},
  \bibinfo{person}{Solomon Messing}, {and} \bibinfo{person}{Sean Westwood}.}
  \bibinfo{year}{2011}\natexlab{}.
\newblock \showarticletitle{Of course {I} wouldn't do that in real life:
  {a}dvancing the arguments for increasing realism in HCI experiments}.
\newblock In \bibinfo{booktitle}{\emph{Extended Abstracts of the CHI Conference
  on Human Factors in Computing Systems}}. \bibinfo{pages}{419--428}.
\newblock


\bibitem[Li et~al\mbox{.}(2019)]%
        {Li.2019}
\bibfield{author}{\bibinfo{person}{Xueni Li}, \bibinfo{person}{Kimmy~Wa Chan},
  {and} \bibinfo{person}{Sara Kim}.} \bibinfo{year}{2019}\natexlab{}.
\newblock \showarticletitle{Service with emoticons: {H}ow customers interpret
  employee use of emoticons in online service encounters}.
\newblock \bibinfo{journal}{\emph{Journal of Consumer Research}}
  \bibinfo{volume}{45}, \bibinfo{number}{5} (\bibinfo{year}{2019}),
  \bibinfo{pages}{973--987}.
\newblock


\bibitem[Lim et~al\mbox{.}(2019)]%
        {Lim.2019}
\bibfield{author}{\bibinfo{person}{Kwan~Hui Lim}, \bibinfo{person}{Jeffrey
  Chan}, \bibinfo{person}{Shanika Karunasekera}, {and}
  \bibinfo{person}{Christopher Leckie}.} \bibinfo{year}{2019}\natexlab{}.
\newblock \showarticletitle{Tour recommendation and trip planning using
  location-based social media: A survey}.
\newblock \bibinfo{journal}{\emph{Knowledge and Information Systems}}
  \bibinfo{volume}{60} (\bibinfo{year}{2019}), \bibinfo{pages}{1247--1275}.
\newblock


\bibitem[Lutz et~al\mbox{.}(2024)]%
        {Lutz.2024}
\bibfield{author}{\bibinfo{person}{Bernhard Lutz}, \bibinfo{person}{Marc Adam},
  \bibinfo{person}{Stefan Feuerriegel}, \bibinfo{person}{Nicolas
  Pr{\"o}llochs}, {and} \bibinfo{person}{Dirk Neumann}.}
  \bibinfo{year}{2024}\natexlab{}.
\newblock \showarticletitle{Which linguistic cues make people fall for fake
  news? {A} comparison of cognitive and affective processing}.
\newblock \bibinfo{journal}{\emph{Proceedings of the ACM on Human-Computer
  Interaction}} \bibinfo{volume}{8}, \bibinfo{number}{CSCW1}
  (\bibinfo{year}{2024}), \bibinfo{pages}{1--22}.
\newblock


\bibitem[Mann and Whitney(1947)]%
        {Mann.1947}
\bibfield{author}{\bibinfo{person}{Henry~B Mann} {and}
  \bibinfo{person}{Donald~R Whitney}.} \bibinfo{year}{1947}\natexlab{}.
\newblock \showarticletitle{On a test of whether one of two random variables is
  stochastically larger than the other}.
\newblock \bibinfo{journal}{\emph{The Annals of Mathematical Statistics}}
  (\bibinfo{year}{1947}), \bibinfo{pages}{50--60}.
\newblock


\bibitem[Okoso et~al\mbox{.}(2025)]%
        {Okoso.2025}
\bibfield{author}{\bibinfo{person}{Ayano Okoso}, \bibinfo{person}{Mingzhe
  Yang}, {and} \bibinfo{person}{Yukino Baba}.} \bibinfo{year}{2025}\natexlab{}.
\newblock \showarticletitle{Do Expressions Change Decisions? {E}xploring the
  Impact of {AI}'s Explanation Tone on Decision-Making}. In
  \bibinfo{booktitle}{\emph{Proceedings of the 2025 CHI Conference on Human
  Factors in Computing Systems}}. \bibinfo{pages}{1--22}.
\newblock


\bibitem[Packard and Berger(2021)]%
        {Packard.2021}
\bibfield{author}{\bibinfo{person}{Grant Packard} {and} \bibinfo{person}{Jonah
  Berger}.} \bibinfo{year}{2021}\natexlab{}.
\newblock \showarticletitle{How concrete language shapes customer
  satisfaction}.
\newblock \bibinfo{journal}{\emph{Journal of Consumer Research}}
  \bibinfo{volume}{47}, \bibinfo{number}{5} (\bibinfo{year}{2021}),
  \bibinfo{pages}{787--806}.
\newblock


\bibitem[Pennebaker et~al\mbox{.}(2015)]%
        {Pennebaker.2015}
\bibfield{author}{\bibinfo{person}{James~W. Pennebaker},
  \bibinfo{person}{Ryan~L. Boyd}, \bibinfo{person}{Kayla Jordan}, {and}
  \bibinfo{person}{Kate Blackburn}.} \bibinfo{year}{2015}\natexlab{}.
\newblock \bibinfo{booktitle}{\emph{The Development and Psychometric Properties
  of LIWC2015}}.
\newblock \bibinfo{publisher}{LIWC.net}, \bibinfo{address}{Austin, TX}.
\newblock


\bibitem[Pugh(2001)]%
        {Pugh.2001}
\bibfield{author}{\bibinfo{person}{S~Douglas Pugh}.}
  \bibinfo{year}{2001}\natexlab{}.
\newblock \showarticletitle{Service with a smile: Emotional contagion in the
  service encounter}.
\newblock \bibinfo{journal}{\emph{Academy of Management Journal}}
  \bibinfo{volume}{44}, \bibinfo{number}{5} (\bibinfo{year}{2001}),
  \bibinfo{pages}{1018--1027}.
\newblock


\bibitem[Roy et~al\mbox{.}(2011)]%
        {Roy.2011}
\bibfield{author}{\bibinfo{person}{Senjuti~Basu Roy}, \bibinfo{person}{Gautam
  Das}, \bibinfo{person}{Sihem Amer-Yahia}, {and} \bibinfo{person}{Cong Yu}.}
  \bibinfo{year}{2011}\natexlab{}.
\newblock \showarticletitle{Interactive itinerary planning}. In
  \bibinfo{booktitle}{\emph{2011 IEEE 27th International Conference on Data
  Engineering}}. IEEE, \bibinfo{pages}{15--26}.
\newblock


\bibitem[Seymour et~al\mbox{.}(2025)]%
        {Seymour.2024}
\bibfield{author}{\bibinfo{person}{Mike Seymour}, \bibinfo{person}{Lingyao
  Yuan}, \bibinfo{person}{Kai Riemer}, {and} \bibinfo{person}{Alan~R Dennis}.}
  \bibinfo{year}{2025}\natexlab{}.
\newblock \showarticletitle{Less artificial, more intelligent: {U}nderstanding
  affinity, trustworthiness, and preference for digital humans}.
\newblock \bibinfo{journal}{\emph{Information Systems Research}}
  \bibinfo{volume}{36}, \bibinfo{number}{2} (\bibinfo{year}{2025}),
  \bibinfo{pages}{1096--1128}.
\newblock


\bibitem[Song et~al\mbox{.}(2025)]%
        {Song.2025}
\bibfield{author}{\bibinfo{person}{Xinheng Song}, \bibinfo{person}{Haiwen
  Song}, \bibinfo{person}{Bingjie Gao}, \bibinfo{person}{Qijun Gao},
  \bibinfo{person}{Wenting Li}, \bibinfo{person}{Linci Xu}, {and}
  \bibinfo{person}{Zhaolin Lu}.} \bibinfo{year}{2025}\natexlab{}.
\newblock \showarticletitle{“Do You Need the Sage's Tea or the Friend's
  Cola” Exploring the Differential Healing Effects of Generative AI
  Conversational Styles}. In \bibinfo{booktitle}{\emph{Extended Abstracts of
  the CHI Conference on Human Factors in Computing Systems}}.
  \bibinfo{pages}{1--8}.
\newblock


\bibitem[Townsend et~al\mbox{.}(2013)]%
        {Townsend.2013}
\bibfield{author}{\bibinfo{person}{Zac Townsend}, \bibinfo{person}{Jack
  Buckley}, \bibinfo{person}{Masataka Harada}, {and} \bibinfo{person}{Marc~A
  Scott}.} \bibinfo{year}{2013}\natexlab{}.
\newblock \showarticletitle{The choice between fixed and random effects}.
\newblock \bibinfo{journal}{\emph{The SAGE Handbook of Multilevel Modeling}}
  (\bibinfo{year}{2013}), \bibinfo{pages}{73--88}.
\newblock


\bibitem[Udandarao et~al\mbox{.}(2025)]%
        {Udandarao.2025}
\bibfield{author}{\bibinfo{person}{Vikranth Udandarao},
  \bibinfo{person}{Noel~Abraham Tiju}, \bibinfo{person}{Muthuraj Vairamuthu},
  \bibinfo{person}{Harsh Mistry}, {and} \bibinfo{person}{Dhruv Kumar}.}
  \bibinfo{year}{2025}\natexlab{}.
\newblock \bibinfo{title}{Roamify: Designing and Evaluating an LLM Based Google
  Chrome Extension for Personalised Itinerary Planning}.
\newblock
\showeprint[arxiv]{2504.10489}~[cs.HC]
\urldef\tempurl%
\url{https://arxiv.org/abs/2504.10489}
\showURL{%
\tempurl}


\bibitem[Wang et~al\mbox{.}(2025)]%
        {Shen.2025}
\bibfield{author}{\bibinfo{person}{Kaimin Wang}, \bibinfo{person}{Yuanzhe
  Shen}, \bibinfo{person}{Changze Lv}, \bibinfo{person}{Xiaoqing Zheng}, {and}
  \bibinfo{person}{Xuan-Jing Huang}.} \bibinfo{year}{2025}\natexlab{}.
\newblock \showarticletitle{TripTailor: A Real-World Benchmark for Personalized
  Travel Planning}. In \bibinfo{booktitle}{\emph{Findings of the Association
  for Computational Linguistics: ACL 2025}}. \bibinfo{pages}{9705--9723}.
\newblock


\bibitem[Wang et~al\mbox{.}(2022)]%
        {Wang.2022}
\bibfield{author}{\bibinfo{person}{Yan Wang}, \bibinfo{person}{Wei Song},
  \bibinfo{person}{Wei Tao}, \bibinfo{person}{Antonio Liotta},
  \bibinfo{person}{Dawei Yang}, \bibinfo{person}{Xinlei Li},
  \bibinfo{person}{Shuyong Gao}, \bibinfo{person}{Yixuan Sun},
  \bibinfo{person}{Weifeng Ge}, \bibinfo{person}{Wei Zhang}, {et~al\mbox{.}}}
  \bibinfo{year}{2022}\natexlab{}.
\newblock \showarticletitle{A systematic review on affective computing:
  {E}motion models, databases, and recent advances}.
\newblock \bibinfo{journal}{\emph{Information Fusion}}  \bibinfo{volume}{83}
  (\bibinfo{year}{2022}), \bibinfo{pages}{19--52}.
\newblock


\bibitem[Wang et~al\mbox{.}(2024)]%
        {Wang.2024}
\bibfield{author}{\bibinfo{person}{Zhiyuan Wang}, \bibinfo{person}{Varun
  Reddy}, \bibinfo{person}{Karen Ingersoll}, \bibinfo{person}{Tabor
  Flickinger}, {and} \bibinfo{person}{Laura~E Barnes}.}
  \bibinfo{year}{2024}\natexlab{}.
\newblock \showarticletitle{Rapport matters: {E}nhancing {HIV} mhealth
  communication through linguistic analysis and large language models}. In
  \bibinfo{booktitle}{\emph{Extended Abstracts of the CHI Conference on Human
  Factors in Computing Systems}}. \bibinfo{pages}{1--8}.
\newblock


\bibitem[Xie et~al\mbox{.}(2023)]%
        {Xie.2023}
\bibfield{author}{\bibinfo{person}{Jian Xie}, \bibinfo{person}{Yidan Liang},
  \bibinfo{person}{Jingping Liu}, \bibinfo{person}{Yanghua Xiao},
  \bibinfo{person}{Baohua Wu}, {and} \bibinfo{person}{Shenghua Ni}.}
  \bibinfo{year}{2023}\natexlab{}.
\newblock \showarticletitle{{QUERT}: Continual pre-training of language model
  for query understanding in travel domain search}. In
  \bibinfo{booktitle}{\emph{Proceedings of the 29th ACM SIGKDD Conference on
  Knowledge Discovery and Data Mining}}. \bibinfo{pages}{5282--5291}.
\newblock


\bibitem[Xie et~al\mbox{.}(2024b)]%
        {Xie.2024.Preprint}
\bibfield{author}{\bibinfo{person}{Jian Xie}, \bibinfo{person}{Kai Zhang},
  \bibinfo{person}{Jiangjie Chen}, \bibinfo{person}{Tinghui Zhu},
  \bibinfo{person}{Renze Lou}, \bibinfo{person}{Yuandong Tian},
  \bibinfo{person}{Yanghua Xiao}, {and} \bibinfo{person}{Yu Su}.}
  \bibinfo{year}{2024}\natexlab{b}.
\newblock \showarticletitle{TravelPlanner: {A} Benchmark for Real-World
  Planning with Language Agents}. In \bibinfo{booktitle}{\emph{ICLR 2024
  Workshop on Large Language Model (LLM) Agents}}.
\newblock


\bibitem[Xie et~al\mbox{.}(2024a)]%
        {Xie.2024}
\bibfield{author}{\bibinfo{person}{Yuguang Xie}, \bibinfo{person}{Changyong
  Liang}, \bibinfo{person}{Peiyu Zhou}, {and} \bibinfo{person}{Li Jiang}.}
  \bibinfo{year}{2024}\natexlab{a}.
\newblock \showarticletitle{Exploring the influence mechanism of
  chatbot-expressed humor on service satisfaction in online customer service}.
\newblock \bibinfo{journal}{\emph{Journal of Retailing and Consumer Services}}
  \bibinfo{volume}{76} (\bibinfo{year}{2024}), \bibinfo{pages}{103599}.
\newblock


\bibitem[Zhang et~al\mbox{.}(2016)]%
        {zhang.2016.Trips}
\bibfield{author}{\bibinfo{person}{Chenyi Zhang}, \bibinfo{person}{Hongwei
  Liang}, {and} \bibinfo{person}{Ke Wang}.} \bibinfo{year}{2016}\natexlab{}.
\newblock \showarticletitle{Trip recommendation meets real-world constraints:
  {POI} availability, diversity, and traveling time uncertainty}.
\newblock \bibinfo{journal}{\emph{ACM Transactions on Information Systems}}
  \bibinfo{volume}{35}, \bibinfo{number}{1} (\bibinfo{year}{2016}),
  \bibinfo{pages}{1--28}.
\newblock


\bibitem[Zhang et~al\mbox{.}(2024)]%
        {Zhang.2024}
\bibfield{author}{\bibinfo{person}{Junbo Zhang}, \bibinfo{person}{Qi Chen},
  \bibinfo{person}{Jiandong Lu}, \bibinfo{person}{Xiaolei Wang},
  \bibinfo{person}{Luning Liu}, {and} \bibinfo{person}{Yuqiang Feng}.}
  \bibinfo{year}{2024}\natexlab{}.
\newblock \showarticletitle{Emotional expression by artificial intelligence
  chatbots to improve customer satisfaction: Underlying mechanism and boundary
  conditions}.
\newblock \bibinfo{journal}{\emph{Tourism Management}}  \bibinfo{volume}{100}
  (\bibinfo{year}{2024}), \bibinfo{pages}{104835}.
\newblock


\bibitem[Zhang and Sun(2016)]%
        {Zhang.2016.CHI}
\bibfield{author}{\bibinfo{person}{Lanyun Zhang} {and} \bibinfo{person}{Xu
  Sun}.} \bibinfo{year}{2016}\natexlab{}.
\newblock \showarticletitle{Designing a trip planner application for groups:
  exploring group tourists? Trip planning requirements}. In
  \bibinfo{booktitle}{\emph{Extended Abstracts of the CHI Conference on Human
  Factors in Computing Systems}}. \bibinfo{pages}{1329--1336}.
\newblock


\bibitem[Zhu et~al\mbox{.}(2025)]%
        {Zhu.2025}
\bibfield{author}{\bibinfo{person}{Shengqi Zhu}, \bibinfo{person}{Jeffrey~M
  Rzeszotarski}, {and} \bibinfo{person}{David Mimno}.}
  \bibinfo{year}{2025}\natexlab{}.
\newblock \showarticletitle{Data paradigms in the era of llms: On the
  opportunities and challenges of qualitative data in the wild}. In
  \bibinfo{booktitle}{\emph{Extended Abstracts of the CHI Conference on Human
  Factors in Computing Systems}}. \bibinfo{pages}{1--8}.
\newblock


\end{thebibliography}


\newpage
\appendix

\section{More Example Conversations}
\subsubsection*{Group A}

\begin{userbubble}\footnotesize
\userlabel{A}: \emph{``Make me an itinerary for exploring NTUs campus. Don't give timings, just give me an overall time. It should start at canteen 2 and include a stop at hall 8''}
\end{userbubble}

\begin{aibubble}\footnotesize
\ailabel: Here's an exciting itinerary for exploring NTU's campus, starting at Canteen 2 and including a visit to Hall 8:
\begin{enumerate}[label=\textbf{\arabic*.}]\itemsep0.2em
\item \textbf{Begin at Canteen 2:} Start your exploration with a delightful meal or coffee at Canteen 2 to fuel your adventure.

\item \textbf{Walk through the Academic Spine:} Wander through the heart of the campus, admiring the innovative architecture of academic buildings.

\item \textbf{Visit the Hive:} Explore this iconic building known for its unique shape and eco-friendly design.

\item \textbf{Check out the Lee Kong Chian School of Medicine:} If you're interested in science, marvel at the modern facilities and learn about its programs.

\item \textbf{Explore the Arts \& Media Precinct:} If you're a culture enthusiast, spend time exploring the School of Art, Design \& Media with its lush rooftop garden.

\item \textbf{Tour Hall of Residence 8:} Stop by Hall 8, known for its vibrant student community and activities, and get a sense of campus life.

\item \textbf{Relax at Yunnan Garden:} End your exploration at Yunnan Garden, a serene spot perfect for reflection and soaking in the beauty of NTU.
\end{enumerate}
Overall, this itinerary should take you about 3-4 hours, allowing you to discover the vibrant academic environment and architectural wonders of NTU! Enjoy the journey! \emoji{herb} \emoji{school}
\end{aibubble}

\subsubsection*{Group B}
\begin{userbubble}\footnotesize
\userlabel{B}: \emph{``what are the visa requirements for indians with a US visa''}
\end{userbubble}
\begin{aibubble}\footnotesize
\ailabel: Indian citizens with a valid U.S. visa can visit Albania without an Albanian visa for stays up to 90 days within a 180-day period. It is important to ensure that the U.S. visa is valid throughout your stay in Albania. Additionally, always check for the latest regulations or contact an Albanian consulate since visa policies can change.
\end{aibubble}

\subsubsection*{Group C}

\begin{userbubble}\footnotesize
\userlabel{C}: \emph{``Change the entire itinerary to be a foodie trip''}
\end{userbubble}
\begin{aibubble}\footnotesize
\ailabel: Here is a food-focused itinerary for your trip to Austin:\\
\textbf{Day 1: Austin's Culinary Beginnings}
\begin{itemize}\itemsep0.2em
    \item \textbf{Morning}: Start your day with breakfast at 'Paperboy', known for its fresh ingredients and creative dishes. It's a perfect breakfast spot for those who love hearty and traditional flavors.
    \item \textbf{Afternoon}: Grab lunch at 'Franklin Barbecue', a legendary spot for indulgent and decadent Texas barbecue. Be ready for a line, but the mouth-watering brisket is worth the wait.
    \item \textbf{Evening}: Dine at 'Odd Duck', which offers a menu featuring locally sourced ingredients and innovative dishes that cater to both vegetarians and meat-lovers. End your night with a dessert at 'Amy's Ice Creams', famous for its playful flavors and quirky vibe.
\end{itemize}  
\textbf{Day 2: Gourmet Discovery}
\begin{itemize}\itemsep0.2em
    \item \textbf{Morning}: Enjoy breakfast at 'Bouldin Creek Cafe', a favorite for its vegetarian-friendly options and hearty breakfast.
    \item \textbf{Afternoon}: Savor lunch at 'Lenoir', a charming spot with a seasonally changing menu that emphasizes local ingredients and eclectic flavors.
    \item \textbf{Evening}: Experience 'Uchi', an upscale sushi restaurant renowned for its creative Japanese dishes. Finish the evening with cocktails at 'The Roosevelt Room', a stylish bar with a vast selection of craft cocktails.
\end{itemize} 
\textbf{Day 3: Tasty Treasures}
\begin{itemize}\itemsep0.2em
    \item \textbf{Morning}: Have breakfast at 'Kerbey Lane Cafe', popular for its delicious pancakes and diverse menu.
    \item \textbf{Afternoon}: For lunch, visit 'The Picnic', a food truck park offering diverse food options, ensuring a delightful culinary adventure.
    \item \textbf{Evening}: End with dinner at 'Fonda San Miguel', celebrated for its authentic Mexican cuisine in a beautiful setting. Enjoy a final treat with live music and drinks at 'The Continental Club', enjoying the classic Austin nightlife.
\end{itemize}   
This itinerary emphasizes traditional, indulgent, and hearty culinary experiences while including spots with a quirky and urban modern aesthetic.
\end{aibubble}

\end{document}